\documentclass[11pt]{article}
\usepackage{blindtext}

\usepackage[abbrvbib, preprint]{jmlr2e}
\usepackage{lastpage}
\jmlrheading{TBD}{2022}{1-\pageref{LastPage}}{1/21}{TBD}{TBD}{Jahan C. Penny-Dimri}

\usepackage{url}
\usepackage{hyperref}
\usepackage[utf8]{inputenc}
\usepackage[small]{caption}
\usepackage{graphicx}
\usepackage{amsmath}
\usepackage{amsthm}
\usepackage{booktabs}
\usepackage{algorithm2e}
\usepackage{amsthm}
\usepackage{tabularx}
\usepackage{booktabs}
\usepackage{multirow}
\usepackage{longtable}
\usepackage{subcaption}
\usepackage{chngpage}
\usepackage{siunitx}
\usepackage{float}
\usepackage{cleveref}
\usepackage{pdflscape}

\newtheorem{proposition}{Proposition}
\newtheorem{remark}{Remark}
\newtheorem{definition}{Definition}

\IfFileExists{upquote.sty}{\usepackage{upquote}}{}
\IfFileExists{microtype.sty}{
  \usepackage[]{microtype}
  \UseMicrotypeSet[protrusion]{basicmath} 
}{}
\usepackage{xcolor}
\newlength{\cslhangindent}
\newlength{\csllabelwidth}
\newlength{\cslentryspacingunit} 
\newenvironment{CSLReferences}[2] 
 {
  \setlength{\parindent}{0pt}
  \ifodd #1
  \let\oldpar\par
  \def\par{\hangindent=\cslhangindent\oldpar}
  \fi
  \setlength{\parskip}{#2\cslentryspacingunit}
 }%
 {}
\usepackage{calc}

\providecommand{\tightlist}{%
  \setlength{\itemsep}{0pt}\setlength{\parskip}{0pt}}

\begin{document}

\title{Dealing with missing data using attention and latent space
regularization}

\author{\name Jahan C.
Penny-Dimri\thanks{Corresponding Author}   \email jahan.penny-dimri@monash.edu \\
\addr Department of Surgery \\
Monash University \\
Melbourne, Australia \\
\AND
\name Christoph Bergmeir  \email christoph.bergmeir@monash.edu \\
\addr Department of Data Science and AI \\
Monash University \\
Melbourne, Australia \\
\AND
\name Julian Smith  \email julian.smith@monash.edu \\
\addr Department of Surgery \\
Monash University \\
Melbourne, Australia \\
}

\RestyleAlgo{ruled}
\SetKwComment{Comment}{/* }{ */}

\maketitle

\begin{abstract}
Most practical data science problems encounter missing data. A wide
variety of solutions exist, each with strengths and weaknesses that
depend upon the missingness-generating process. Here we develop a
theoretical framework for training and inference using only observed
variables enabling modeling of incomplete datasets without imputation.
Using an information and measure-theoretic argument we construct models
with latent space representations that regularize against the potential
bias introduced by missing data. The theoretical properties of this
approach are demonstrated empirically using a synthetic dataset. The
performance of this approach is tested on 11 benchmarking datasets with
missingness and 18 datasets corrupted across three missingness patterns
with comparison against a state-of-the-art model and industry-standard
imputation. We show that our proposed method overcomes the weaknesses of
imputation methods and outperforms the current state-of-the-art.
\end{abstract}

\hypertarget{introduction}{%
\section{Introduction}\label{introduction}}

Missing data is a common problem encountered by the data scientist. The
consequences of missing data are often not straightforward and depend on
the missingness generating process (Little and Rubin 2002). Choosing the
best strategy to deal with missingness is critical when designing
statistical or machine learning models that rely on incomplete datasets.
A frequent complication is that the missingness generating process is
often unknown, leading to assumptions about the missingness and
potentially the introduction of bias into model training and inference
(Davey and Dai 2020). The two most commonly employed strategies for
dealing with missing data are to either drop data points where missing
data exists or impute the values (Bertsimas et al. 2021). Both
strategies can potentially introduce bias into a model if applied
incorrectly (Little and Rubin 2002).

The `impute and regress' strategy has been recently critiqued in the
setting of machine learning predictors with the finding that an
imputation method leads to a consistent prediction model if that model
can almost surely undo the imputation, which questions the need for
sophisticated imputation strategies (Bertsimas et al. 2018, 2021). The
best case scenario for an imputation method is to replace missing
unobserved variables with values that do not corrupt the model's
predictive distribution from the true predictive distribution given only
the observed variables. The worst case would be introducing a
significant bias during training that increases the divergence of the
model's predictive distribution from the true distribution. Indeed,
Jeong et al. (2022) proved with an information-theoretic argument that
there is no universally fair imputation method for different downstream
learning tasks.

While practitioners continue to develop novel machine learning and
statistical methods for imputation, relatively less effort has been made
to create a framework to reason about models that can fit incomplete
data with the notable exception of decision tree models (Gavankar and
Sawarkar 2015; Jeong et al. 2022). In this paper, we consider the
question of designing a model that trains and performs inference on only
the observed variables, without imputation or data deletion.

\hypertarget{related-work}{%
\subsection{Related Work}\label{related-work}}

Most recent work has focused on developing novel imputation techniques,
such as auto-encoders, in order to build better imputations based on the
structure of the data (Abiri et al. 2019). A consideration of learning
and inference without imputation has been considered recently by
Bertsimas et al. (2021) who provided a theoretical approach to the
deficiencies of imputation in predictive modeling, drawing an important
distinction between statistical inference and prediction. They build on
this approach to develop an adaptive hierarchical linear regression
model that is capable of performing prediction in the presence of any
missingness pattern. Jeong et al. (2022) provide a method of inference
without imputation using a decision tree approach, with a
fairness-regularized loss function. Our work differs substantially from
these approaches, whereby we show that training and inference using only
the observed data is feasible using latent space representations and an
entropy-based objective which ensures regularization against the
potential bias of missingness. An important contribution utilizing
latent representations is the partial variational auto-encoder (Ma et
al. 2019). This approach was developed for inference in recommendation
systems where there is a partially observed vector. Our work extends
this prior approach to build a theoretical basis for utilizing latent
representations, and demonstrates empirically the properties of such a
model.

\hypertarget{summary-of-contributions}{%
\subsection{Summary of contributions}\label{summary-of-contributions}}

\begin{enumerate}
\def\labelenumi{\arabic{enumi}.}
\tightlist
\item
  Introduce a novel method of dealing with missingness by establishing a
  simple framework for reasoning about a model that fits and infers from
  the observed variables only.
\item
  Interpret the latent space representations in this framework with a
  measure- and information theoretic argument.
\item
  Empirically validate the theoretical properties of the latent space on
  a synthetic dataset with a latent space attention model.
\item
  Demonstrate the effectiveness of this model on benchmark datasets with
  corrupted data and real world datasets with missingness.
\end{enumerate}

\hypertarget{interpreting-missingness-in-a-measurable-space}{%
\section{Interpreting missingness in a measurable
space}\label{interpreting-missingness-in-a-measurable-space}}

We first describe a sample space \(\Omega\) that is defined by an
unknown data generating process and unknown missingness generating
process. We then define three random variables:
\[ X_{d}: \Omega \mapsto \mathbb{R} \]
\[ Y: \Omega \mapsto \mathbb{R} \] \[ M_{d}: \Omega \mapsto \{0,1\} \]

Where \(d \in \{1, ..., D\}\) is the total number of possibly observed
variables. We can further specify, \(X = \{X_{1}, ..., X_{D}\}\), which
is the random variable \(X: \Omega \mapsto \mathbb{R}^{D}\), and
\(M = \{M_{1}, ..., M_{D}\}\), which is the random variable
\(M: \Omega \mapsto \{0,1\}^{D}\). A realization of \(X\) is a vector
representation of possibly observed variables and \(Y\) is the outcome
variable of interest. \(M\) is a missingness vector where a value of 1
indicates a missing value and 0 an observed value.

If we consider the power set \(\mathcal{U} = \mathcal{P}(X)\), then we
can define a measurable space \((X, \mathcal{U})\). The impact of
missingness results in a smaller \(\sigma\)-algebra such that
\(U = \{i: i \in \mathcal{U} \land j \in \mathcal{M} \land 1 \notin j \}\),
where \(\mathcal{M} = \mathcal{P}(M)\). Finally, we can define a
probability space \((X, U, \mu)\) where a measure maps each combination
of possible variables in \(X\), defined in \(U\), to a probability
whereby \(\mu: u \mapsto [0,1]\), where \(u \in U\).

An implication of our definition of the measure, \(\mu\), is that where
no missingness exists, each subset is equally likely, and the
distribution is uniform. In the presence of a missingness generating
process, some subsets are more likely than others and the distribution
diverges from uniform.

How the probability distribution diverges from uniform can be considered
with reference to the definitions of missingness originally described by
Little and Rubin (2002). If
\(X_{obs} = \{x: x \in X \land m \in M \land m = 0 \}\) and
\(X_{mis} = \{x: x \in X \land m \in M \land m = 1 \}\) and we consider
\(M\) to be defined by some unknown parameter \(\beta\), with a
conditional distribution \(f(M|X,\beta)\), then we can define three
types of missingness:

\begin{definition} \label{def_mcar}

If \(f(M|X,\beta)\) = \(f(M|\beta)\) then M \(\perp X|\beta\) and data
is considered to be missing completely at random (MCAR)

\end{definition}

\begin{definition} \label{def_mar}

If \(f(M|X,\beta)\) = \(f(M|X_{obs},\beta)\) then M
\(\perp X_{mis}|X_{obs},\beta\) and data is considered to be missing at
random (MAR)

\end{definition}

\begin{definition} \label{def_mnar}

If \(f(M|X,\beta) \ne f(M|\beta)\) and
\(f(M|X,\beta) \ne f(M|X_{obs},\beta)\), then data is considered to be
missing not at random (MNAR)

\end{definition}

We can extend these definitions to define conditional versions of
\(\mu\) with \(\mu_{U|X}\) and \(\mu_{U|X_{obs}}\). If
\(\mu(U) = \mu_{U|X}(U|X)\) then missing data can be considered MCAR. If
\(\mu_{U|X}(U|X) = \mu_{U|X_{obs}}(U|X_{obs})\) then missing data can be
considered MAR. Finally, if
\(\mu(U) \ne \mu_{U|X}(U|X) \ne \mu_{U|X_{obs}}(U|X_{obs})\) then data
is MNAR.

From this initial description of missingness, we can consider a naïve
first attempt at predicting \(Y\) from \(X\) in the presence of
missingness.

\hypertarget{using-an-ensemble-to-learn-the-power-set-of-features}{%
\section{Using an ensemble to learn the power set of
features}\label{using-an-ensemble-to-learn-the-power-set-of-features}}

As a starting point to developing our proposed method, we first consider
a simple ensemble approach. We construct a hypothetical ensemble of
\(K\) models, parameterized by \(\theta_k \in \Theta\), which performs
the mapping for each combination of variables to our outcome space,
\(f_{\theta_k}: U_k \mapsto Y\), where \(U_k \in U\),
\(k \in \{1, ... K\}\) and \(K = |U|\). If the input vector \(x \sim X\)
contains missing values then only the complete subsets in \(u \sim U\)
with no missing values are used to update the models. To perform
inference, we could simply choose the output of the function with the
highest cardinality set. This strategy presupposes that all input data
contains information related to the outcome, and feature selection has
already occurred prior to model creation. In this setting, this model
definition allows us to make a prediction, \(\hat{y}\), given any subset
of available variables from \(X\).

This approach is only valid for datasets with missingness defined as
MCAR as there are two implicit strategies occurring. The first is simply
excluding the variable with missing data from our model by defining our
subsets in \(U\) which exclude that variable. This is always a valid
strategy as it carries no risk of introducing bias into the remaining
data, at the cost of losing information associated with that variable.
The second implicit strategy is that in subsets that include variables
with missing values, we remove all items in that subset that have a
missing value. This is akin to dropping rows in a tabular dataset. By
dropping missing data that is either MAR or MNAR, bias is introduced
into the remaining data and the resultant model that is fitted to that
data will incorporate that bias. In the ``ensemble of \(K\)'' models
described above, models fitted on subsets with missing data will
incorporate the biases of a dropping data strategy.

\hypertarget{incorporating-missingness-in-a-latent-space-to-overcome-bias}{%
\section{Incorporating missingness in a latent space to overcome
bias}\label{incorporating-missingness-in-a-latent-space-to-overcome-bias}}

In order to overcome the deficiencies of the previous formulation we
could try to introduce a latent space representation for \(U\) in our
model and attempt to incorporate missingness into this latent space. We
redefine \(f_{\theta_k}\) to map each input vector in \(U\) to a latent
space \(Z\), \(f_{\theta_k}: U_k \mapsto Z\), which is shared across all
possible subsets and define a second model, parameterized by \(\Phi\)
that maps Z to our output space, \(g_{\Phi}: Z \mapsto Y\). We have now
defined an ``ensemble of compositions'',
\(g_{\Phi} \circ f_{\theta_k}\).

Previous work by Boudiaf et al. (2020) has derived the result that when
a latent space model is fit with a cross-entropy objective,
\(\mathcal{H}(Y;\hat{Y}|\hat{Z})\), this is equivalent to maximizing the
mutual information \(I(\hat{Z};Y)\).

Using this result, we can reason about the representations of subsets in
the latent space. Consider two subsets \(U_i, U_j \in U\), which have
identical variables, but one has an additional variable
\(|U_{i} - U_{j}| = 1\). If both \(U_i\) and \(U_j\) contain the same
amount of information with respect to the outcome, then they will be
clustered to the same location in latent space.

To formalize this clustering effect in terms of a measurable space, we
can define a measure space for our dataset
\((\mathcal{D},\mathcal{A},\mathcal{I})\), such that
\(\mathcal{D} = \{ \{x, y\}: x \in X \land y \in Y\}\),
\(\mathcal{A} = \{ \{u, y\}: u \in U \land y \in Y\}\), and
\(\mathcal{I}\) is a measure based on mutual information defined as
\(\mathcal{I} = I(f(U_{k}); Y)\), where \(f\) is a general function that
maps a subset, \(U_k \in U\) to the latent space \(Z\).

Mutual information is a correct basis for a measure in this measurable
space as it satisfies the properties of non-negativity, mapping to zero
for the empty set, and countable additivity across the subsets defined
in \(U\). The property of countable additivity relies on the
independence of their latent space representations of \(U_k\). This
independence arises in two ways. Firstly, although we do not guarantee
\(X_i \perp X_j\) for \(X_i, X_j \in X\) there is an independence across
subsets of \(X\) based on the construction of \(U\), such that
\(U_i \perp U_j\) for \(U_i, U_j \in U\). The independence of the
subsets \(U_k\) guarantees the independence of their latent space
representations. Secondly, we can notice that \(\mathcal{I}\) is really
measuring the mutual information of \(Z|U_k\), and in the conditional
universe of the latent space \(Z|U_i \perp Z|U_j\).

Using our measure, \(\mathcal{I}\), and model definition we can formally
define the effect of the mutual information on the latent space.

\begin{proposition} \label{prop:equality}

If \(\mathcal{I}(U_{i}; Y) = \mathcal{I}(U_{j}; Y)\) where
\(i,j \in \{1,..., K\}\) and \(|U_{i} - U_{j}| = 1\), then
\(f(U_{i}) = f(U_{j}) = \hat{Z} | U_{i} = \hat{Z} | U_{j}\).

\end{proposition}

We can broaden this equality with a geometric interpretation.

\begin{proposition} \label{prop:distance}

\(| \mathcal{I}(U_{i}; Y) - \mathcal{I}(U_{j}; Y)| \propto d(f(U_{i}), f(U_{j}))\)
where \(i,j \in \{1,..., K\}\), \(|U_{i} - U_{j}| = 1\), and \(d\) is a
distance metric in the latent space, such as the Euclidean distance.

\end{proposition}

The effect of Proposition \ref{prop:distance} is that the latent space
representations for each \(U_k\) in our ``ensemble of compositions''
cluster according to the mutual information. Additionally, there is a
guarantee that the model learns representations for high cardinality
subsets that are regularized to the highest information but lowest
cardinality subset. This obviates the need for explicit feature
selection, as non-informative features are not represented in the latent
space.

\begin{remark} \label{remark:1}

If \(Y\) is a Bernoulli-distributed variable parameterized by the output
of \(g\), \(Y = Bern(g(f_k(U_k)))\), then in the extreme case where no
input features carry information with respect to our outcome all subsets
will cluster to the latent space representation of the empty set, which
is mapped to the point of highest entropy in \(Y\).

\end{remark}

In order to understand how the latent space representation can help with
missingness, we need only consider what happens to
\(\mathcal{I} = I(f(U_{k}); Y)\) as missingness affects \(U_{k}\). If an
unknown missingness generating process affects \(U_k\), then
\(\mathcal{I} = I(f(U_{k}); Y)\) will decrease. This will be true
whether the missingness is MCAR, MAR or MNAR. This reduction is directly
represented in the latent space. When missingness is present in \(U_k\),
the representation in \(Z\) is then regularized to the lowest
cardinality and highest information subset.

Finally, it is interesting to note that there are many cases where
missingness generating processes of the previously described patterns
may also provide information related to the outcome (Li et al. 2018).
This is commonly encountered, for example, in healthcare datasets where
healthy people often have missing data because there was no indication
to perform a test. We can define this property as arising when
\(I(M_d; Y) > 0\). In our latent space model,
\(\mathcal{I} = I(f(U_{k}); Y)\) for a subset \(U_k\) which is affected
by informative missingness may actually increase.

\hypertarget{our-approach-using-attention-to-model-the-latent-space}{%
\section{Our approach: using attention to model the latent
space}\label{our-approach-using-attention-to-model-the-latent-space}}

There is an obvious limitation to the ensemble approach, which is poor
computational scaling as the number of models in the ensemble, \(K\),
increases exponentially with the dimensions of the dataset. Rather than
defining a set of functions \(f_{\theta_k}\) for each
\(k \in {1, ..., K}\), we can define a general function \(f_{\theta}\),
which can map each \(U_k\) to the latent space, as we did in
Propositions \ref{prop:equality} and \ref{prop:distance}. A natural
choice for such a model that can take heterogeneous length inputs is to
adapt an attention based model normally used to solve sequence tasks.

Scaled dot product attention has been used to construct state-of-the-art
sequence models such as the Transformer (Vaswani et al. 2017). The
Transformer maps an input sequence \((x_{1},..., x_{d})\) to a latent
space representation \((z_{1},..., z_{d})\), from which a decoder
outputs a sequence of \((y_{1},..., y_{m})\). Here each
\(x, y, z \in \mathbb{R}^{\text{embed}}\), which is the embedding
dimensionality. The similarity to the aforementioned ``ensemble of
compositions'' can begin to be appreciated at this point. To approximate
the ensemble with a Transformer architecture we must add two components.
Firstly we add a feature specific embedding network that maps our input
vector to some higher dimensional embedding space,
\(f_{d}: \mathbb{R} \mapsto \mathbb{R}^{\text{embed}}\). Then we apply a
Transformer-style model where the output is instead
\(z \in \mathbb{R}^{\text{embed}}\), which is then mapped to the output
space,
\(g_{\phi}: \mathbb{R}^{\text{embed}} \mapsto \mathbb{R}^{\text{out}}\).
In this model, the output of the Transformer model
\(z \in \mathbb{R}^{\text{embed}}\), is analogous to the latent space in
the previously compositional model and the function \(g_{\phi}\) is
serving the same purpose in both models.

It is important to note that while the compositional ensemble explicitly
trains a separate model for each subset, the Transformer must train on
randomly generated subsets at each training step. This can be achieved
using dropout to stochastically create a mask for the input features
applied at each attention module. Concrete dropout has been shown to
enable feature ranking and applying it at the feature level as a subset
sampling process means that more informative features are included in
the subsets more frequently (Chang et al. 2018). This choice of model
design helps further regularize the output of the model toward the
lowest cardinality, highest information subset.

Finally, we introduce our method as a latent space attention model
(LSAM), which can deal with missingness `out-of-the-box'. In summary,
this approach uses a neural network to embed each feature, then maps the
set of available embeddings to a latent space using a Transformer-style
network. It is the latent space representation which is regularized
against missingness as it trains. The latent space representation is
then mapped to an outcome space using a neural network. The pseudocode
for the training procedure is demonstrated in Algorithm
\ref{fig:algorithm} and training details can be found in the Appendix B.

\begin{algorithm}
Initialize network parameters $\rho, \psi, \theta, \phi$\;
\For{epoch \text{in} Epochs}{
  \For{$x, y$ in batched($X \in \mathbb{R}^{n \times d}, Y \in \mathbb{R}^{n \times out}$)}{
			$mask \sim Concrete(\rho)$ \Comment*[r]{Sampling feature mask for subsetting}
			$e \gets f(\psi, x)$ \Comment*[r]{Embedding network $f_\psi: \mathbb{R}^{d} \to \mathbb{R}^{d \times \text{embed}}$}
			$z \gets t(\theta, e, mask)$ \Comment*[r]{Attention network $t_\theta: \mathbb{R}^{d \times \text{embed}} \to \mathbb{R}^{\text{embed}}$}
			$\hat{y} \gets g(\phi, z)$ \Comment*[r]{Output network $g_\phi: \mathbb{R}^{\text{embed}} \to \mathbb{R}^{\text{out}}$}
			$g \gets \nabla_{\rho, \psi, \theta, \phi}l(y, \hat{y})$ \Comment*[r]{Gradient of cross-entropy loss}
			$\rho, \psi, \theta, \phi \gets update(g, \rho, \psi, \theta, \phi)$
	}
}
\caption{Latent space attention model training procedure.
  A stochastically generated mask, drawn from a concrete distribution at each training step, ensures the model learns from different subsets of variables.}\label{fig:algorithm}
\end{algorithm}

\hypertarget{experiments}{%
\section{Experiments}\label{experiments}}

Models are built using JAX and trained using adaptive stochastic
gradient optimization with early stopping (Kingma and Ba 2017; Bradbury
et al. 2018; Zhuang et al. 2020).

Descriptions of hyperparameter optimization and model training can be
found in Appendix B. Code for the model and experiments is available at
\url{https://github.com/jahanpd/Missingness}.

\hypertarget{model-characteristics}{%
\subsection{Model Characteristics}\label{model-characteristics}}

We first explore the properties of our proposed model with regard to the
previously defined latent space using a toy dataset. Our aim is to show
the effect of both the mutual information of a predictor on the latent
space representations, and the effect of missingness on these
representations.

\hypertarget{latent-space-representation}{%
\subsubsection{Latent space
representation}\label{latent-space-representation}}

In order to explore the representation of input variables in the latent
space, we use the classic 2-dimensional synthetic spiral dataset. This
is a non-linear binary classification problem with two input variables,
\(x1\) and \(x2\), representing the x and y axes respectively. The 2
dimensional dataset is augmented with 2 additional variables. The first,
\(x3\), is random noise from a standard Gaussian carrying no information
of the outcome, \(x3 \sim \mathcal{N}(0,1)\). The second, \(x4\), is the
outcome variable corrupted with random noise from a uniform distribution
and therefore carrying some information about the outcome,
\(x4 = y + \epsilon\), \(\epsilon \sim \mathcal{U}_{[0,1]}\).

Both the LSAM and compositional ensemble model are trained on the
4-dimensional dataset. Outcome measures are then bootstrapped with
experiments repeated 30 times. We can obtain a statistical measure of
distance between groups reported using a Student t-test and measure the
\(p\)-value.

\begin{table}[h]
\begin{center}
\begin{tabular}{crrrr} 
\toprule
    & \multicolumn{2}{c}{LSAM} & \multicolumn{2}{c}{Ensemble} \\
\cmidrule(lr){2-3} \cmidrule(lr){4-5}
$u_{k}$ & \{\} & $p$-value & \{\} & $p$-value \\ [0.5ex] 
\midrule
  \midrule
\{x1\}          &  3.69 &  \num{3.95e-10} &     0.29 &  \num{4.43e-05} \\
\{x2\}          &  4.18 &  \num{2.16e-11} &     0.38 &  \num{3.40e-04} \\
\{x1, x2\}      &  5.10 &  \num{4.13e-14} &     2.74 &  \num{6.78e-05} \\
\{x3 (noise)\}  &  3.44 &  \num{1.14e-08} &     0.04 &  \num{1.04e-03} \\
\{x4 (signal)\} &  5.15 &  \num{3.77e-14} &     4.37 &  \num{6.08e-05} \\
\bottomrule
\end{tabular}
\end{center}
\caption{Euclidean distance between latent space representations of $u_{k}$ and the empty set
for the LSAM and Ensemble based models. The $p$-value represents the distributional difference.
The difference is greatest where the signal is strongest, and least where there is only random noise.}
\label{tbl:table1}
\end{table}

We first explore the Euclidean distance for various subsets compared to
the empty set. From Propositions \ref{prop:equality} and
\ref{prop:distance}, we would predict that the distance is shorter when
the subset is only random noise, \(x3\), compared to carrying some
information about the outcome. Table \ref{tbl:table1} shows that this is
indeed the case, with the distance between the empty set and noise being
the smallest value for both the ensemble and the LSAM. The statistical
difference between the bootstrapped distributions is much less for the
non-informative noise compared to subsets carrying information.

We can then explore the Euclidean distance between \(u_{k}\), where
\(u_{k} \subseteq \{x1, x2\}\), and \(u_{k}\) + \(x\), where
\(x \subset \{x3,x4\}\). From Propositions \ref{prop:equality} and
\ref{prop:distance}, we predict that the distance is shorter when \(x\)
is random noise, as in \(x3\), compared to carrying information about
the outcome, as in \(x4\). This is confirmed in Table \ref{tbl:table2}
where we see that adding noise changes the location of the mapping in
latent space less than adding an informative variable across both
models.

\begin{table*}[h]
\begin{center}
\begin{tabular}{c c c c c c c} 
\toprule
    & \multicolumn{3}{c}{LSAM} & \multicolumn{3}{c}{Ensemble} \\
\cmidrule(lr){2-4} \cmidrule(lr){5-7}
$u_{k}$ & +\{noise\} & +\{signal\} & $p$-value & +\{noise\} & +\{signal\} & $p$-value \\
\midrule
\midrule
\{x1\}     &     0.43 &      4.16 &  \num{3.09e-12} &     0.16 &      2.41 &  \num{8.24e-05} \\
\{x2\}     &     0.37 &      3.18 &  \num{1.02e-11} &     0.14 &      3.92 &  \num{1.80e-04} \\
\{x1, x2\} &     0.21 &      1.09 &  \num{3.77e-07} &     2.63 &      3.79 &  \num{1.43e-02} \\
\{\}       &     3.44 &      5.15 &  \num{2.02e-07} &     0.04 &      4.37 &  \num{6.28e-05} \\
\bottomrule
\end{tabular}
\end{center}
\caption{Euclidean distance between latent space representations of $u_{k}$ and $u_{k}$ + either a noise
variable or an informative variable for both a LSAM or Ensemble based model.
The location in latent space moves the most when an informative variable is added to a subset.}
\label{tbl:table2}
\end{table*}

\hypertarget{missingness-regularization}{%
\subsubsection{Missingness
regularization}\label{missingness-regularization}}

We now explore the effect of the level of missingness on the latent
space representation. In order to do this we measure the Euclidean
distance between \(u_{k}\), where \(u_{k} \subseteq \{x1, x2\}\), and
\(u_{k}\) + \(\{x4\}\), where the degree of missingness is increased in
\(x4\) from 0\% of the data to 99\% of the data. From Proposition
\ref{prop:distance}, we would predict that as missingness in \(x4\)
increases, the distance between \(u_{k}\) and \(u_{k}\) + \(\{x4\}\)
decreases. Table \ref{tbl:table3} empirically demonstrates this effect
where the location in latent space converges to approximately the same
location by 99\% missingness.

\begin{table}[h]
\begin{center}
\setlength{\tabcolsep}{3pt}
\begin{tabular}{crrrrrr} 
\toprule
$u_{k}$ & 0\% & 20\% & 40\% & 60\% & 80\% & 99\% \\
\midrule
\midrule
\{x1\}     &  7.84 &  5.93 &  4.62 &  2.88 &  1.48 &  0.07 \\
\{x2\}     &  6.11 &  4.42 &  3.40 &  2.24 &  1.05 &  0.05 \\
\{x1, x2\} &  2.07 &  0.79 &  0.57 &  0.23 &  0.13 &  0.01 \\
\{\}       &  9.13 &  6.88 &  5.42 &  3.45 &  1.77 &  0.09 \\
\bottomrule
\end{tabular}

\end{center}
\caption{Euclidean distance between latent space representations of $u_{k}$ and $u_{k}$ +
an informative variable ($x4$) for varying levels of missingness in the informative variable ($x4$).
The location of the subset in latent space moves less when the missingness in the added informative variable increases.}
\label{tbl:table3}
\end{table}

\hypertarget{missingness-feature-importance-and-concrete-dropout}{%
\subsubsection{Missingness, feature importance and concrete
dropout}\label{missingness-feature-importance-and-concrete-dropout}}

Finally, we explore the effect of missingness on the concrete
distribution learned during training. Specifically, we can look at how
the probability of dropping \(x4\) from subsets during training changes
as the level of missingness increases from 0\% to 99\% of the variable.
Table \ref{tbl:table4} shows that the probability of dropout increases
toward 0.5, the point of maximum entropy, as missingness increases to
99\%. The implication of this finding is that the model assigns more
importance to variables with less missing data, even when the variable
with missing data is an informative variable.

\begin{table}[h]
\begin{center}
\begin{tabular}{c c c c c c c} 
\toprule
Variable & 0\% & 20\% & 40\% & 60\% & 80\% & 99\% \\
\midrule
\midrule
x1     &  0.34 &  0.33 &  0.32 &  0.33 &  0.33 &  0.34 \\
x2     &  0.32 &  0.32 &  0.32 &  0.32 &  0.31 &  0.33 \\
x3 (noise) &  0.50 &  0.49 &  0.49 &  0.49 &  0.48 &  0.49 \\
x4 (signal)       &  0.40 &  0.38 &  0.37 &  0.38 &  0.40 &  0.46 \\
\bottomrule
\end{tabular}
\end{center}
  \caption{Learned probabilities, from a concrete distribution, of dropping variables during training across different levels of
    missingness in the informative variable ($x4$). The probability of dropping $x4$ from training approaches that of random noise ($x3$)
    as the missingness increases in that variable.}
\label{tbl:table4}
\end{table}

\hypertarget{benchmarking-performance}{%
\subsection{Benchmarking Performance}\label{benchmarking-performance}}

The OpenML-CC18 classification benchmarking suite is used to assess
performance in two settings (Vanschoren et al. 2013).

\hypertarget{dataset-selection}{%
\subsubsection{Dataset selection}\label{dataset-selection}}

In the first set of experiments, we only select datasets with complete
data availability, so we can controllably corrupt them with specific
missingness patterns. In the second set of experiments, we select
datasets with missingness, however the missingness pattern is unknown.
For all experiments, we select tabular datasets with a maximum number of
250 features and rows between 1000 and 10000 for computability. For
datasets with pre-existing missingness we included datasets with greater
than or equal to 5\% of the data missing. Our final collection includes
18 complete datasets and 11 datasets with missingness. Details on the
methods used to corrupt the datasets with missingness are available in
the Appendix A.

\hypertarget{comparator-models}{%
\subsubsection{Comparator models}\label{comparator-models}}

In all experiments, we compare `out-of-the-box' performance with
missingness against industry standard data processing pipelines with
imputation. We compare our LSAM approach against a high performing
ensemble decision tree model, LightGBM, which is also capable of dealing
with missing data natively (Ke et al. 2017).

Imputation can be broadly grouped into simple imputation strategies or
regression methods. Simple imputation can be the replacement of a
missing value by a statistic, a pre-specified value, or encoding the
missingness as a new variable. Examples include mean or mode imputation.
This strategy can lead to consistent predictions in the setting of MCAR,
but is not appropriate where data is MAR or MNAR (Bertsimas et al.
2021). Despite the strong assumption needed, the benefit to this
approach is computational efficiency and ease of implementation.

Regression strategies are often combined with multiple imputation
techniques, which involves regressing multiple times with many models in
order to generate a sample distribution of imputed values (Rubin 1987;
Buuren and Groothuis-Oudshoorn 2011). Multiple imputation can be
performed with linear regression, decision tree, or even random forest
models and can provide unbiased estimates in the setting of data MCAR
and MAR (Stekhoven and Buhlmann 2011). Although this approach can
require significant computational resources, especially in the setting
of large datasets, it has become the gold standard for most modeling
tasks that require imputation. In our experiments, we compare native
model performance against 3 imputation strategies: simple imputation
with mode or mean imputation based on data type, an iterative
multivariate imputation strategy, and multiple imputation with random
forests (MiceForest) (Pedregosa et al. 2011; Stekhoven and Buhlmann
2011).

\hypertarget{performance-on-complete-and-corrupted-real-world-datasets}{%
\subsubsection{Performance on complete and corrupted real-world
datasets}\label{performance-on-complete-and-corrupted-real-world-datasets}}

We first report the performance on the baseline complete datasets. For
the metric of accuracy, the LSAM performs better with 13/18 wins. For
the metric of negative log-likelihood (NLL) the LSAM is better with
12/18 wins.

\begin{figure}
    \centering
    \subcaptionbox{Performance when data is MCAR}
        {\includegraphics[width=0.7\textwidth]{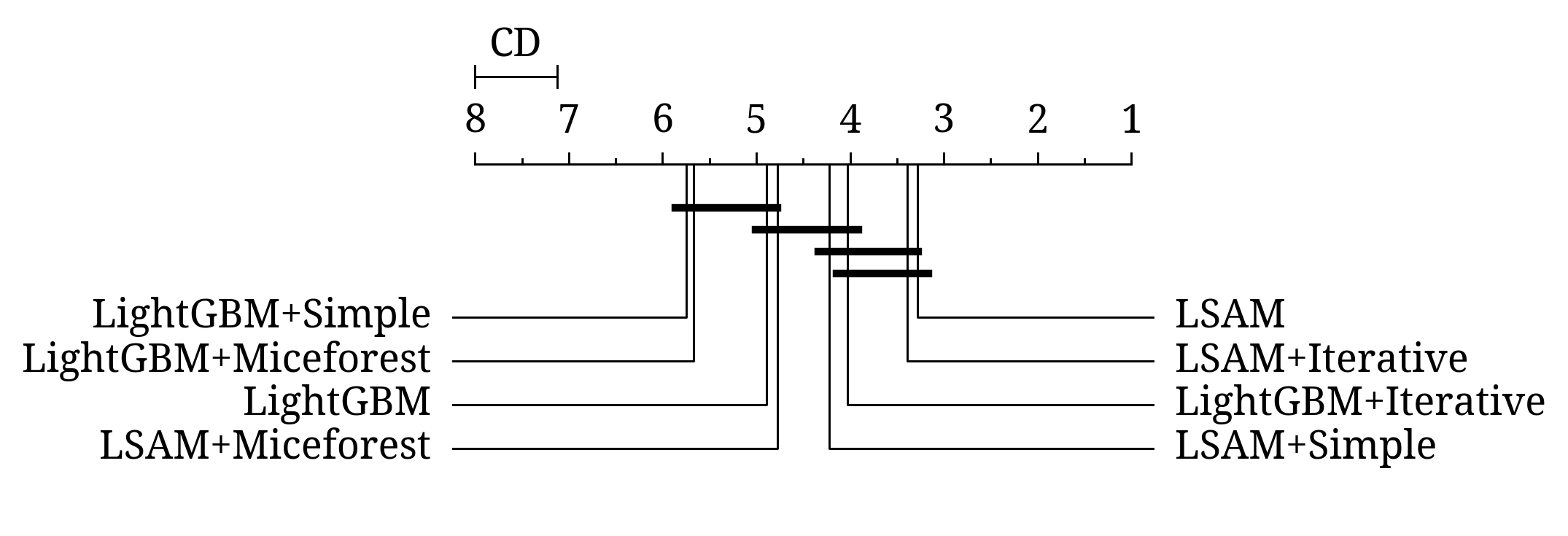}}%
    \vspace{0cm}
    \subcaptionbox{Performance when data is MAR}
        {\includegraphics[width=0.7\textwidth]{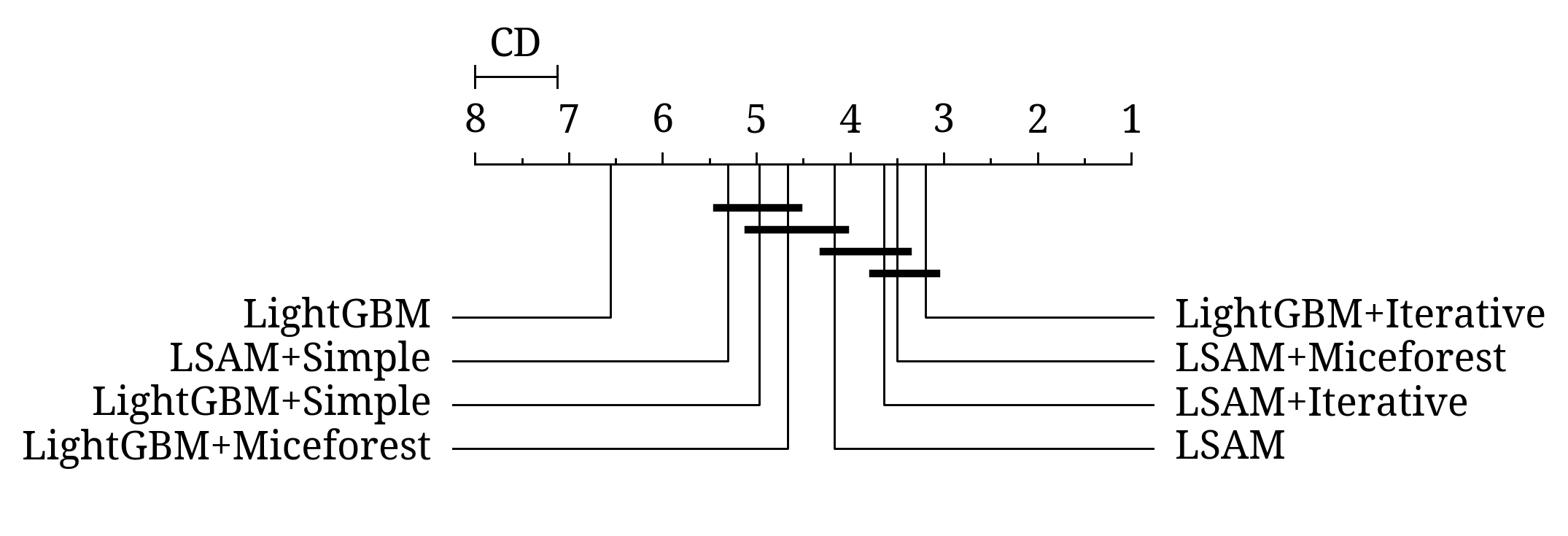}}%

    \subcaptionbox{Performance when data is MNAR}
        {\includegraphics[width=0.7\textwidth]{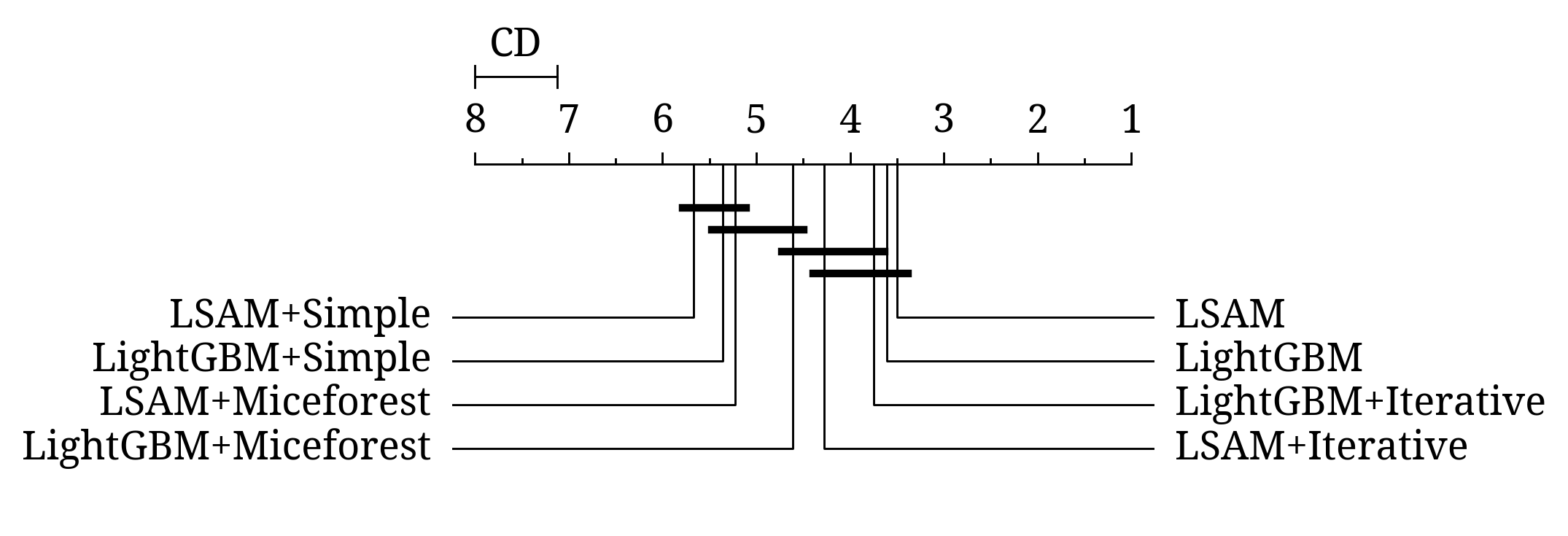}}

    \caption{
    Critical difference diagrams comparing performance for different missingness regimes, demonstrating improved performance for the LSAM without imputation.
    Points are labelled by the type of model as well as the imputation strategy if used.
    The performance metric is the change in negative log-likelihood from baseline performance with complete data. Further right in the diagram indicates better performance.
    A break in the solid bar underneath demonstrates statistical significance.
    }
    \label{fig:figure1}
\end{figure}

\begin{figure}[h]
    \centering
    \begin{subfigure}[b]{1.0\textwidth}
        \centering
        \caption{Performance when data is MCAR}
        \includegraphics[scale=0.49]{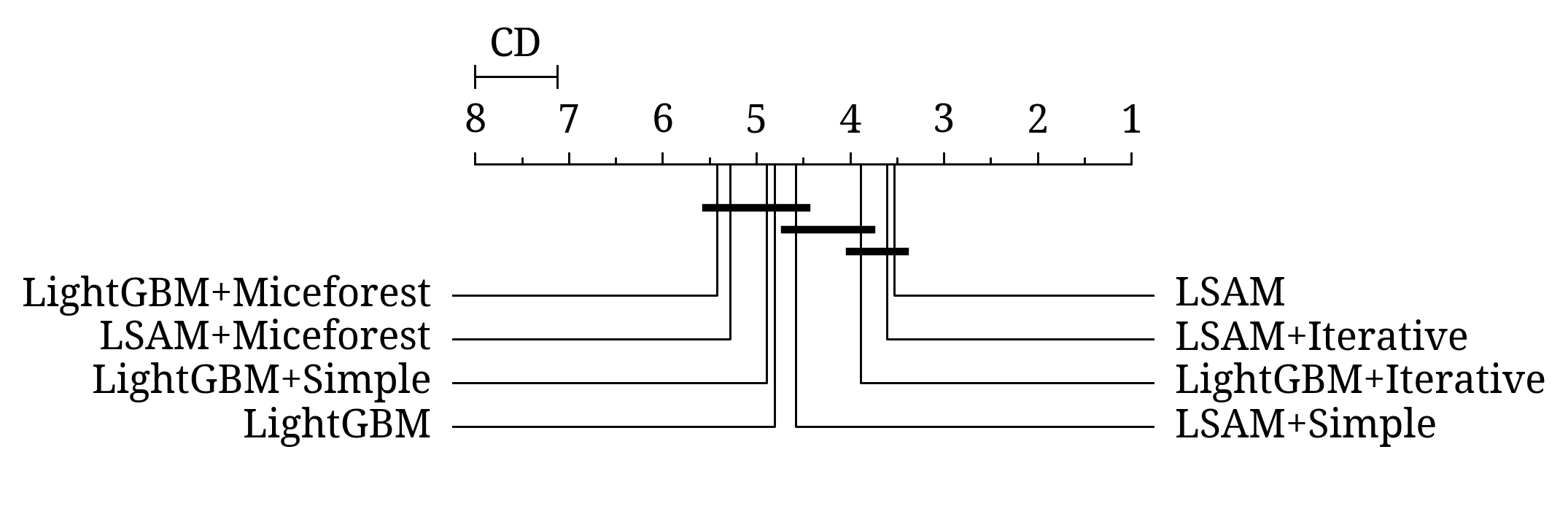}
    \end{subfigure}

    \begin{subfigure}[b]{1.0\textwidth}
        \centering
        \caption{Performance when data is MAR}
        \includegraphics[scale=0.49]{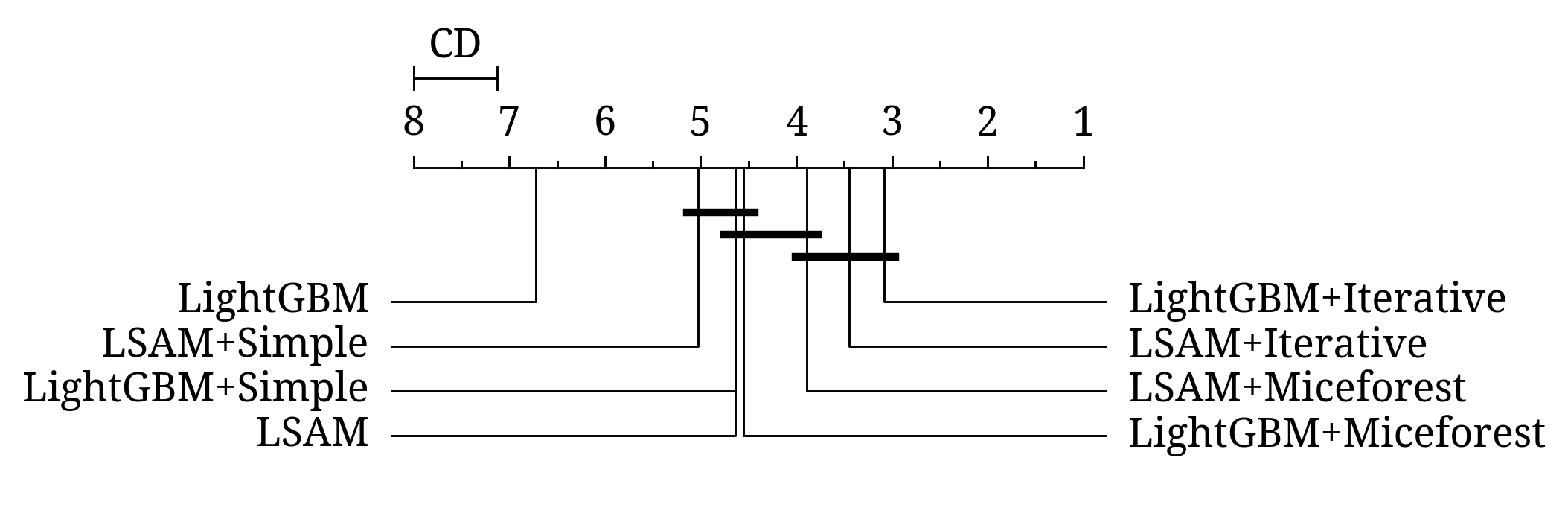}
    \end{subfigure}

    \begin{subfigure}[b]{1.0\textwidth}
        \centering
        \caption{Performance when data is MNAR}
        \includegraphics[scale=0.49]{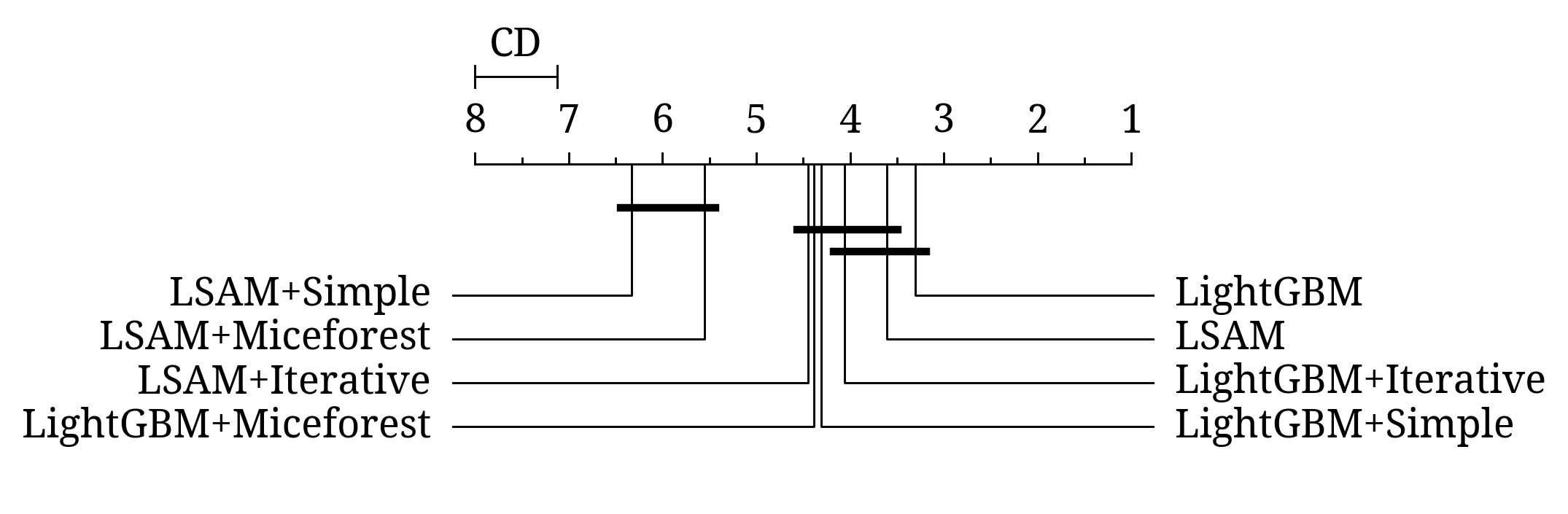}
    \end{subfigure}

    \caption{Critical difference diagrams comparing performance for different missingness regimes, demonstrating improved performance for the LSAM without imputation.
    Points are labelled by the type of model as well as the imputation strategy if used.
    The performance metric is the change in accuracy from baseline performance with complete data. Further right in the diagram indicates better performance.
    A break in the solid bar underneath demonstrates statistical significance.
    }
    \label{fig:figure3}
\end{figure}

To statistically compare the performance of all possible approaches we
use critical difference diagrams with a Nemenyi two-tailed test (Demšar
2006). Figures \ref{fig:figure1} and \ref{fig:figure3} shows these plots
in the context of the negative log-likelihood and accuracy respectively.
These demonstrate that the LSAM, using out-of-the-box performance,
outperforms or performs as well as an `impute and regress' strategy when
data is MCAR or MNAR. Additionally, the LSAM outperforms LightGBM in
most settings, with the exception of data MAR. Complete tables of
performance results are available in Appendix C.

\hypertarget{performance-on-real-world-datasets-with-missingness}{%
\subsubsection{Performance on real-world datasets with
missingness}\label{performance-on-real-world-datasets-with-missingness}}

\begin{figure}[h]
    \centering
    \subcaptionbox{Performa for the outcome of negative log-likelihood}
        {\includegraphics[width=0.7\textwidth]{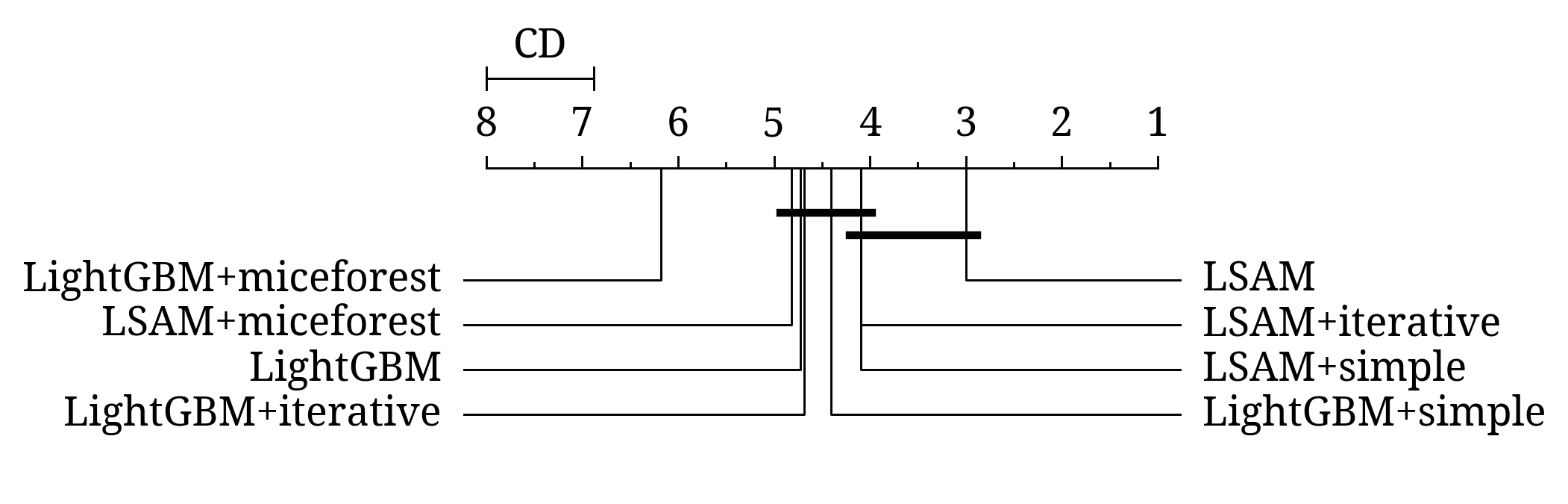}}%
    \vspace{0cm}
    \subcaptionbox{Performance for the outcome of accuracy}
        {\includegraphics[width=0.7\textwidth]{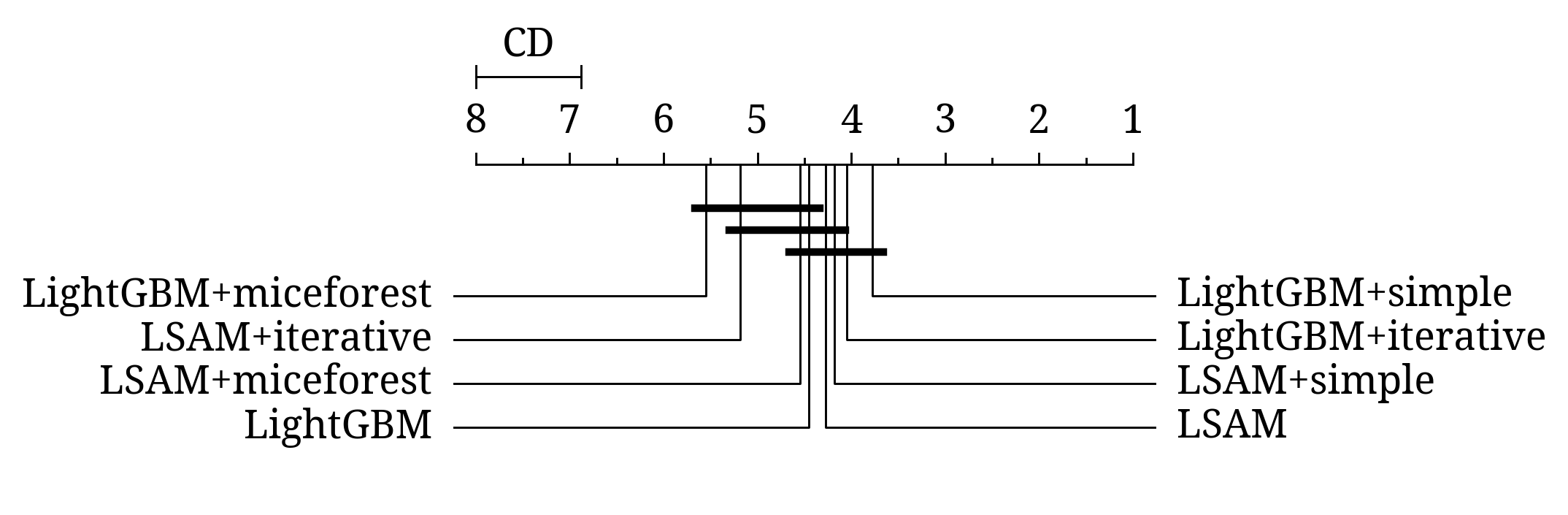}}%
    \caption{Critical difference diagram comparing performance with negative log-likelihood and accuracy on datasets with unknown missingness pattern, demonstrating improved performance of the LSAM model for negative log-likelihood.}
    \label{fig:figure2}
\end{figure}

Again, we compare the results of all possible approaches using a
critical difference diagram. Figure \ref{fig:figure2} shows these plots
for both metrics of accuracy and negative log-likelihood. Performance as
measured by the negative log-likelihood showed a higher performance of
the LSAM using out-of-the-box performance. Performance as measured by
the accuracy showed a higher performance of LightGBM models. These
results indicate that the LSAM model is better calibrated at the expense
of accuracy.

\hypertarget{meta-learning}{%
\subsubsection{Meta-learning}\label{meta-learning}}

Finally, to explore the effect of dataset characteristics when LSAM
outperforms the comparators, we train a random forest model to predict
an LSAM win from the known features of each dataset. Full experiment
details and results are available in the Appendix D. Datasets with
higher dimensionality, a greater number of rows, and a greater
proportion of numeric variables tended to favor LSAM.

\hypertarget{conclusion}{%
\section{Conclusion}\label{conclusion}}

To the best of our knowledge, our work is the first description and
theoretical justification for utilizing an attention based latent space
model for dealing with missingness in tabular datasets. Importantly, we
have contributed a measure and information theoretic argument that
extends upon previous work on using latent space representations to
model partially observed datasets. Based on this argument, we have shown
that regularization occurs in the latent space both theoretically and
empirically using a LSAM and compositional ensemble architecture. The
regularization acts to cluster representations in the latent space
toward the lowest cardinality and highest mutually informative subset of
variables. We have shown this approach to outperform imputation and
LightGBM.

Our results suggest that when data is MCAR and MAR, there is less of a
clear advantage of avoiding imputation. Importantly, when data is MNAR,
our experiments suggest a potential advantage of out-of-the-box methods
to avoid the bias of imputation methods in this setting. This finding is
predicted from our theoretical derivation of latent space
regularization. Performance on unknown, and likely mixed, real-world
missingness patterns corroborates the finding that latent space
regularization can avoid the bias in imputation methods. Interestingly,
LightGBM also out-competes imputation when data is MNAR which likely
arises from gradient based decision trees allocating a split with
missingness to the direction that minimizes the loss (Ke et al. 2017).
We suggest that as data MNAR contains some information about the
underlying value it is likely that splitting based on loss minimization
allows LightGBM to incorporate information from data MNAR.

A limitation of our approach is that it bundles missingness with model
design. Certain predictive tasks with known missingness that is MAR or
MCAR, or specific model requirements, would likely benefit from an
impute and regress strategy instead of our proposed method.
Additionally, while we have introduced latent space regularization to
compositional ensemble and attention-based models, there is no limit to
possible architectures that could achieve the same effect. Future work
should further validate this latent space regularization with other
model designs and explore the application of this approach in other real
world dataset domains, such as imaging or time-series data.

We conclude that for predictive tasks, attention based latent space
models with concrete dropout may be a principled model choice in the
context of missing data with unknown missingness pattern. They require
no assumptions about the missingness pattern of the data and require
minimal extra computational cost compared to a similarly parameterized
model.

\newpage

\acks{There are no sources of funding, financial disclosures or conflicts of interests to declare.}

\newpage

\appendix

\hypertarget{appendix-a.}{%
\section*{Appendix A.}\label{appendix-a.}}
\addcontentsline{toc}{section}{Appendix A.}

For corrupting data with missingness, we have three predefined patterns.

\hypertarget{mcar}{%
\subsection*{MCAR}\label{mcar}}
\addcontentsline{toc}{subsection}{MCAR}

In order to generate data MCAR, 40\% of data from 40\% of randomly
selected columns is randomly deleted.

\hypertarget{mar}{%
\subsection*{MAR}\label{mar}}
\addcontentsline{toc}{subsection}{MAR}

For generating data MAR 40\% of data from 40\% of columns is deleted
such that a value in a row is deleted depending on the value of it's
neighboring column to the left.

\hypertarget{mnar}{%
\subsection*{MNAR}\label{mnar}}
\addcontentsline{toc}{subsection}{MNAR}

Finally, for data MNAR 40\% of data from 40\% of columns is deleted such
that a value in a row is deleted depending on it's value.

\hypertarget{appendix-b.}{%
\section*{Appendix B.}\label{appendix-b.}}
\addcontentsline{toc}{section}{Appendix B.}

\hypertarget{hyper-parameter-optimization}{%
\subsection*{Hyper-parameter
optimization}\label{hyper-parameter-optimization}}
\addcontentsline{toc}{subsection}{Hyper-parameter optimization}

Our approach to optimization was using Bayesian hyperparamater (Biewald
2020). For the transformer model, 4 hyperparameter ranges were optimized
including learning rate (1e-3 - 1e-5), weight decay (1e-1 - 1e-7), model
dimensionality (4 - 150) and optimizer (adam, adabelief, and stochastic
gradient descent).

For the LightGBM model, 5 hyperparameters were optimized including
number of leaves (1 - 1000), the learning rate (1e-1 - 1e-6), the
minimum data per leaf (1 - 500), the maximum number of bins (100 -
1000), and the type of boosting (gbdt, rf, dart, goss).

\hypertarget{model-training}{%
\subsection*{Model training}\label{model-training}}
\addcontentsline{toc}{subsection}{Model training}

Both models were trained using a cross-entropy loss objective. The LSAM
included weight regularization. For LSAM model was trained for a maximum
of 5000 steps with early stopping.

LightGBM models were trained for 100 iterations with early stopping.

\hypertarget{appendix-c.}{%
\section*{Appendix C.}\label{appendix-c.}}
\addcontentsline{toc}{section}{Appendix C.}

Complete results for performance on the benchmark datasets with
difference missingness regimes are available in the following tables.

\begin{table*}[ht]
\caption{Baseline accuracy with no missing data on benchmark datasets.}
\label{tbl:table6}
\centering
\begin{center}
\begin{tabular}{lrr}
    \toprule
      & LSAM &  LightGBM \\
    \midrule
ozone-level-8hr & \textbf{0.92} & 0.90 \\
splice & 0.82 & \textbf{0.92} \\
pc3 & 0.85 & \textbf{0.87} \\
qsar-biodeg & \textbf{0.85} & 0.85 \\
mfeat-zernike & \textbf{0.81} & 0.77 \\
mfeat-fourier & \textbf{0.84} & 0.80 \\
texture & \textbf{0.99} & 0.95 \\
kr-vs-kp & \textbf{0.99} & 0.98 \\
satimage & \textbf{0.91} & 0.90 \\
dna & 0.96 & \textbf{0.96} \\
optdigits & \textbf{0.98} & 0.97 \\
first-order-theorem-proving & 0.50 & \textbf{0.57} \\
GesturePhaseSegmentationProcessed & 0.55 & \textbf{0.66} \\
pc4 & \textbf{0.89} & 0.84 \\
mfeat-karhunen & \textbf{0.97} & 0.95 \\
mfeat-pixel & \textbf{0.97} & 0.85 \\
spambase & \textbf{0.95} & 0.93 \\
mfeat-factors & \textbf{0.97} & 0.95 \\
    \bottomrule
\end{tabular}
\end{center}
\end{table*}

\onecolumn
\begin{landscape}
\begin{table*}[h]
\caption{Change in accuracy from baseline when data is missing completely at random.}
\label{tbl:table7}
\centering
\begin{center}
\setlength\tabcolsep{1.5pt}
\begin{tabular}{lcccccccc}
    \toprule
    & \multicolumn{2}{c}{None} & \multicolumn{2}{c}{Simple} & \multicolumn{2}{c}{Iterative} & \multicolumn{2}{c}{MiceForest} \\\
    & LSAM &  LightGBM & LSAM &  LightGBM & LSAM &  LightGBM & LSAM &  LightGBM \\
    \midrule
ozone-level-8hr & \textbf{-0.01} & -0.01 & -0.01 & -0.01 & -0.01 & -0.02 & -0.01 & -0.02 \\
splice & -0.07 & -0.01 & -0.02 & \textbf{-0.01} & -0.05 & \textbf{-0.01} & -0.04 & -0.02 \\
pc3 & -0.02 & \textbf{-0.00} & -0.02 & -0.00 & -0.01 & -0.01 & -0.02 & -0.01 \\
qsar-biodeg & 0.00 & -0.01 & 0.00 & -0.00 & \textbf{0.00} & -0.00 & 0.00 & -0.00 \\
mfeat-zernike & -0.01 & -0.01 & -0.01 & -0.01 & \textbf{-0.00} & -0.01 & -0.01 & -0.01 \\
mfeat-fourier & \textbf{-0.01} & -0.01 & -0.02 & -0.02 & -0.02 & -0.02 & -0.02 & -0.02 \\
texture & -0.00 & -0.01 & -0.00 & -0.01 & \textbf{-0.00} & -0.01 & -0.00 & -0.01 \\
kr-vs-kp & -0.01 & -0.01 & -0.03 & -0.02 & -0.04 & -0.02 & -0.02 & \textbf{-0.01} \\
satimage & -0.00 & -0.01 & -0.00 & -0.01 & \textbf{0.00} & -0.00 & -0.00 & -0.01 \\
dna & \textbf{0.00} & -0.00 & -0.01 & -0.00 & -0.01 & -0.00 & -0.01 & -0.00 \\
optdigits & -0.00 & -0.01 & \textbf{-0.00} & -0.01 & -0.01 & -0.01 & -0.01 & -0.01 \\
first-order-theorem-proving & -0.01 & -0.02 & -0.01 & -0.02 & \textbf{-0.00} & -0.02 & -0.01 & -0.02 \\
GesturePhaseSegmentationProcessed & -0.02 & -0.03 & -0.02 & -0.03 & \textbf{-0.01} & -0.01 & -0.03 & -0.04 \\
pc4 & -0.00 & 0.01 & -0.00 & 0.01 & -0.00 & \textbf{0.02} & -0.00 & 0.01 \\
mfeat-karhunen & -0.01 & -0.01 & -0.01 & -0.01 & \textbf{-0.00} & -0.01 & -0.01 & -0.01 \\
mfeat-pixel & -0.01 & -0.01 & -0.01 & -0.01 & -0.01 & -0.02 & -0.01 & \textbf{-0.00} \\
spambase & -0.01 & -0.01 & \textbf{-0.00} & -0.01 & -0.01 & -0.01 & -0.01 & -0.01 \\
mfeat-factors & \textbf{-0.00} & -0.00 & -0.01 & -0.00 & -0.01 & -0.00 & -0.01 & -0.01 \\
    \bottomrule
\end{tabular}
\end{center}
\end{table*}
\end{landscape}

\begin{landscape}
\begin{table*}[h]
\caption{Change in accuracy from baseline when data is missing at random.}
\label{tbl:table8}
\centering
\begin{center}
\setlength\tabcolsep{1.5pt}
\begin{tabular}{lcccccccc}
    \toprule
    & \multicolumn{2}{c}{None} & \multicolumn{2}{c}{Simple} & \multicolumn{2}{c}{Iterative} & \multicolumn{2}{c}{MiceForest} \\
    & LSAM &  LightGBM & LSAM &  LightGBM & LSAM &  LightGBM & LSAM &  LightGBM \\
    \midrule
ozone-level-8hr & \textbf{0.00} & -0.01 & -0.00 & -0.00 & -0.01 & -0.01 & -0.00 & -0.01 \\
splice & -0.16 & -0.27 & -0.06 & -0.05 & -0.06 & -0.05 & \textbf{-0.02} & -0.07 \\
pc3 & -0.01 & -0.01 & \textbf{0.00} & -0.00 & -0.01 & 0.00 & -0.00 & -0.01 \\
qsar-biodeg & \textbf{0.01} & -0.01 & -0.00 & -0.01 & 0.01 & -0.01 & -0.00 & -0.02 \\
mfeat-zernike & -0.00 & -0.04 & -0.01 & -0.01 & \textbf{-0.00} & -0.01 & -0.01 & -0.01 \\
mfeat-fourier & -0.02 & -0.03 & -0.02 & -0.01 & -0.01 & -0.01 & -0.02 & \textbf{-0.01} \\
texture & -0.03 & -0.08 & -0.02 & -0.08 & \textbf{-0.00} & -0.02 & -0.01 & -0.07 \\
kr-vs-kp & -0.01 & -0.02 & -0.02 & -0.01 & -0.02 & -0.01 & -0.01 & \textbf{-0.01} \\
satimage & -0.02 & -0.01 & -0.02 & -0.00 & -0.01 & \textbf{-0.00} & -0.01 & -0.00 \\
dna & -0.00 & -0.00 & -0.01 & \textbf{0.00} & -0.01 & \textbf{0.00} & -0.00 & -0.00 \\
optdigits & -0.04 & -0.04 & -0.02 & -0.02 & \textbf{-0.00} & -0.01 & -0.01 & -0.01 \\
first-order-theorem-proving & -0.04 & -0.04 & -0.05 & -0.02 & \textbf{-0.02} & -0.02 & -0.03 & -0.02 \\
GesturePhaseSegmentationProcessed & -0.02 & -0.05 & -0.02 & -0.03 & \textbf{-0.01} & -0.02 & -0.02 & -0.04 \\
pc4 & -0.01 & \textbf{0.02} & -0.01 & 0.02 & -0.02 & 0.02 & 0.00 & 0.01 \\
mfeat-karhunen & -0.03 & -0.09 & -0.02 & -0.02 & -0.01 & \textbf{-0.01} & -0.02 & -0.01 \\
mfeat-pixel & -0.03 & -0.12 & -0.01 & -0.09 & \textbf{-0.01} & -0.07 & -0.01 & -0.09 \\
spambase & -0.01 & -0.01 & -0.01 & -0.02 & \textbf{-0.00} & -0.01 & -0.01 & -0.02 \\
mfeat-factors & \textbf{-0.00} & -0.01 & -0.01 & -0.01 & \textbf{-0.00} & -0.00 & -0.01 & -0.01 \\
    \bottomrule
\end{tabular}
\end{center}
\end{table*}
\end{landscape}

\begin{landscape}
\begin{table*}[h]
\caption{Change in accuracy from baseline when data is missing not at random}
\label{tbl:table9}
\centering
\begin{center}
\setlength\tabcolsep{1.5pt}
\begin{tabular}{lcccccccc}
    \toprule
    & \multicolumn{2}{c}{None} & \multicolumn{2}{c}{Simple} & \multicolumn{2}{c}{Iterative} & \multicolumn{2}{c}{MiceForest} \\
    & LSAM &  LightGBM & LSAM &  LightGBM & LSAM &  LightGBM & LSAM &  LightGBM \\
    \midrule
ozone-level-8hr & -0.00 & -0.00 & -0.01 & -0.01 & -0.01 & -0.02 & -0.02 & \textbf{-0.00} \\
splice & -0.40 & -0.26 & \textbf{-0.10} & -0.12 & -0.14 & -0.12 & -0.27 & -0.23 \\
pc3 & -0.01 & \textbf{0.00} & -0.02 & 0.00 & -0.01 & -0.00 & -0.01 & 0.00 \\
qsar-biodeg & 0.00 & -0.01 & -0.01 & -0.01 & \textbf{0.01} & -0.02 & -0.01 & -0.01 \\
mfeat-zernike & \textbf{-0.01} & -0.01 & -0.04 & -0.02 & -0.02 & -0.01 & -0.03 & -0.02 \\
mfeat-fourier & -0.03 & -0.01 & -0.06 & -0.01 & -0.06 & \textbf{-0.00} & -0.05 & -0.01 \\
texture & -0.03 & -0.04 & -0.02 & -0.03 & \textbf{-0.00} & -0.02 & -0.00 & -0.02 \\
kr-vs-kp & -0.01 & \textbf{-0.01} & -0.05 & -0.04 & -0.05 & -0.04 & -0.02 & -0.02 \\
satimage & -0.01 & -0.01 & -0.03 & -0.01 & -0.02 & \textbf{-0.00} & -0.01 & -0.00 \\
dna & \textbf{0.00} & -0.00 & -0.00 & -0.00 & -0.00 & -0.00 & -0.00 & -0.00 \\
optdigits & -0.00 & \textbf{-0.00} & -0.01 & -0.01 & -0.01 & -0.01 & -0.01 & -0.01 \\
first-order-theorem-proving & -0.04 & -0.06 & -0.10 & -0.05 & \textbf{-0.01} & -0.04 & -0.03 & -0.06 \\
GesturePhaseSegmentationProcessed & \textbf{-0.03} & -0.06 & -0.08 & -0.08 & -0.06 & -0.06 & -0.08 & -0.06 \\
pc4 & -0.01 & -0.01 & -0.01 & 0.00 & -0.02 & 0.01 & -0.02 & \textbf{0.01} \\
mfeat-karhunen & \textbf{-0.02} & -0.02 & -0.07 & -0.03 & -0.05 & -0.02 & -0.05 & -0.02 \\
mfeat-pixel & -0.01 & -0.02 & -0.01 & -0.01 & \textbf{-0.00} & -0.07 & -0.01 & -0.02 \\
spambase & -0.01 & \textbf{-0.00} & -0.02 & -0.01 & -0.01 & -0.01 & -0.01 & -0.01 \\
mfeat-factors & -0.02 & -0.01 & -0.02 & \textbf{-0.01} & -0.01 & -0.01 & -0.02 & -0.01 \\
    \bottomrule
\end{tabular}
\end{center}
\end{table*}
\end{landscape}

\begin{table*}[h]
\caption{Baseline negative log-likelihood with no missing data for the benchmark datasets.}
\label{tbl:table10}
\centering
\begin{center}
\begin{tabular}{lrr}
    \toprule
    & \multicolumn{2}{l}{NLL (No Missing Data)} \\
    & LSAM &  LightGBM \\
    \midrule
ozone-level-8hr & \textbf{0.51} & 0.85 \\
splice & 1.47 & \textbf{0.38} \\
pc3 & 0.92 & \textbf{0.65} \\
qsar-biodeg & 0.72 & \textbf{0.71} \\
mfeat-zernike & \textbf{0.41} & 0.68 \\
mfeat-fourier & \textbf{0.41} & 0.56 \\
texture & \textbf{0.03} & 0.96 \\
kr-vs-kp & \textbf{0.09} & 0.54 \\
satimage & \textbf{0.24} & 0.25 \\
dna & 0.17 & \textbf{0.15} \\
optdigits & \textbf{0.06} & 0.13 \\
first-order-theorem-proving & 1.47 & \textbf{1.25} \\
GesturePhaseSegmentationProcessed & 1.22 & \textbf{0.95} \\
pc4 & \textbf{0.57} & 0.79 \\
mfeat-karhunen & \textbf{0.12} & 0.21 \\
mfeat-pixel & \textbf{0.13} & 2.27 \\
spambase & \textbf{0.31} & 0.44 \\
mfeat-factors & \textbf{0.13} & 0.21 \\
    \bottomrule
\end{tabular}
\end{center}
\end{table*}

\begin{landscape}
\begin{table*}[h]
\caption{Negative change in NLL from baseline when data is missing completely at random (bigger is better).}
\label{tbl:table2}
\centering
\begin{center}
\setlength\tabcolsep{1.5pt}
\begin{tabular}{lcccccccc}
    \toprule
    & \multicolumn{2}{c}{None} & \multicolumn{2}{c}{Simple} & \multicolumn{2}{c}{Iterative} & \multicolumn{2}{c}{MiceForest} \\
    & LSAM &  LightGBM & LSAM &  LightGBM & LSAM &  LightGBM & LSAM &  LightGBM \\
    \midrule
dna & \textbf{0.00} & -0.02 & -0.02 & -0.02 & -0.02 & -0.02 & -0.02 & -0.02 \\
ozone-level-8hr & \textbf{0.00} & -0.02 & -0.01 & -0.02 & -0.01 & -0.03 & -0.03 & -0.03 \\
mfeat-karhunen & -0.01 & -0.03 & -0.02 & -0.04 & \textbf{0.00} & -0.03 & -0.03 & -0.04 \\
GesturePhaseSegmentationProcessed & -0.05 & -0.06 & -0.04 & -0.05 & \textbf{-0.02} & -0.03 & -0.07 & -0.07 \\
optdigits & \textbf{-0.01} & -0.03 & -0.01 & -0.03 & -0.04 & -0.02 & -0.04 & -0.03 \\
mfeat-factors & \textbf{-0.01} & -0.03 & -0.03 & -0.03 & -0.01 & -0.01 & -0.02 & -0.02 \\
mfeat-fourier & \textbf{-0.03} & -0.06 & -0.04 & -0.07 & -0.04 & -0.06 & -0.06 & -0.07 \\
qsar-biodeg & 0.02 & -0.03 & 0.02 & -0.03 & 0.01 & -0.01 & \textbf{0.02} & -0.03 \\
mfeat-zernike & -0.04 & -0.04 & -0.04 & -0.04 & -0.02 & \textbf{-0.02} & -0.03 & -0.04 \\
kr-vs-kp & -0.07 & \textbf{-0.04} & -0.18 & -0.11 & -0.21 & -0.11 & -0.15 & -0.04 \\
pc4 & 0.02 & 0.01 & 0.03 & 0.01 & \textbf{0.06} & 0.01 & 0.05 & 0.01 \\
spambase & -0.03 & -0.03 & -0.03 & -0.04 & \textbf{-0.03} & -0.04 & -0.05 & -0.04 \\
satimage & -0.01 & -0.01 & -0.01 & -0.01 & \textbf{0.00} & 0.00 & -0.01 & -0.01 \\
pc3 & -0.04 & \textbf{-0.01} & -0.04 & -0.02 & -0.04 & -0.02 & -0.04 & -0.02 \\
texture & -0.01 & -0.02 & -0.02 & -0.02 & \textbf{-0.01} & -0.01 & -0.01 & -0.02 \\
mfeat-pixel & -0.03 & -0.00 & -0.03 & -0.00 & -0.02 & -0.00 & -0.04 & \textbf{0.00} \\
splice & -1.28 & -0.02 & \textbf{0.98} & -0.03 & -1.22 & -0.03 & 0.96 & -0.05 \\
first-order-theorem-proving & -0.00 & -0.05 & -0.02 & -0.04 & \textbf{0.00} & -0.04 & -0.01 & -0.04 \\
    \bottomrule
\end{tabular}
\end{center}
\end{table*}
\end{landscape}

\begin{landscape}
\begin{table*}[h]
\caption{Negative change in NLL from baseline when data is missing at random (bigger is better).}
\label{tbl:table2}
\centering
\begin{center}
\setlength\tabcolsep{1.5pt}
\begin{tabular}{lcccccccc}
    \toprule
    & \multicolumn{2}{c}{None} & \multicolumn{2}{c}{Simple} & \multicolumn{2}{c}{Iterative} & \multicolumn{2}{c}{MiceForest} \\
    & LSAM &  LightGBM & LSAM &  LightGBM & LSAM &  LightGBM & LSAM &  LightGBM \\
    \midrule
dna & 0.01 & -0.01 & -0.03 & -0.01 & -0.03 & -0.01 & \textbf{0.01} & -0.01 \\
ozone-level-8hr & \textbf{0.05} & -0.06 & 0.04 & -0.01 & -0.07 & -0.02 & 0.04 & -0.02 \\
mfeat-karhunen & -0.08 & -0.25 & -0.08 & -0.07 & -0.04 & \textbf{-0.04} & -0.05 & -0.06 \\
GesturePhaseSegmentationProcessed & -0.05 & -0.10 & -0.04 & -0.06 & \textbf{-0.03} & -0.05 & -0.05 & -0.09 \\
optdigits & -0.14 & -0.13 & -0.06 & -0.07 & \textbf{-0.02} & -0.03 & -0.05 & -0.05 \\
mfeat-factors & -0.01 & -0.07 & -0.02 & -0.04 & \textbf{-0.00} & -0.02 & -0.02 & -0.04 \\
mfeat-fourier & -0.07 & -0.12 & -0.07 & -0.05 & -0.04 & \textbf{-0.04} & -0.06 & -0.04 \\
qsar-biodeg & \textbf{0.01} & -0.02 & -0.02 & -0.03 & -0.01 & -0.01 & -0.01 & -0.05 \\
mfeat-zernike & -0.04 & -0.15 & -0.06 & -0.06 & \textbf{-0.02} & -0.02 & -0.04 & -0.05 \\
kr-vs-kp & -0.06 & -0.04 & -0.12 & -0.06 & -0.10 & -0.06 & -0.07 & \textbf{-0.03} \\
pc4 & -0.00 & -0.01 & -0.01 & -0.01 & -0.05 & 0.00 & \textbf{0.02} & -0.00 \\
spambase & -0.05 & -0.09 & -0.05 & -0.09 & \textbf{-0.03} & -0.06 & -0.07 & -0.08 \\
satimage & -0.05 & -0.03 & -0.05 & -0.01 & -0.02 & \textbf{-0.00} & -0.01 & -0.01 \\
pc3 & 0.05 & -0.01 & \textbf{0.06} & 0.02 & -0.08 & 0.02 & 0.02 & -0.00 \\
texture & -0.09 & -0.18 & -0.06 & -0.19 & \textbf{-0.01} & -0.05 & -0.02 & -0.16 \\
mfeat-pixel & -0.10 & -0.00 & -0.03 & \textbf{-0.00} & -0.02 & -0.01 & -0.03 & -0.00 \\
splice & 0.50 & -0.53 & -0.14 & -0.14 & -0.13 & -0.14 & \textbf{0.91} & -0.16 \\
first-order-theorem-proving & -0.09 & -0.06 & -0.08 & -0.03 & \textbf{-0.02} & -0.04 & -0.04 & -0.03 \\
    \bottomrule
\end{tabular}
\end{center}
\end{table*}
\end{landscape}

\begin{landscape}
\begin{table*}[h]
\caption{Negative change in NLL from baseline when data is missing not at random (bigger is better).}
\label{tbl:table13}
\centering
\begin{center}
\setlength\tabcolsep{1.5pt}
\begin{tabular}{lcccccccc}
    \toprule
    & \multicolumn{2}{c}{None} & \multicolumn{2}{c}{Simple} & \multicolumn{2}{c}{Iterative} & \multicolumn{2}{c}{MiceForest} \\
    & LSAM &   LightGBM & LSAM &  LightGBM & LSAM &  LightGBM & LSAM &   LightGBM \\
    \midrule
dna & \textbf{0.00} & -0.00 & -0.02 & -0.01 & -0.02 & -0.01 & -0.00 & -0.01 \\
ozone-level-8hr & 0.02 & -0.03 & \textbf{0.08} & -0.03 & -0.02 & -0.03 & 0.03 & -0.02 \\
mfeat-karhunen & \textbf{-0.04} & -0.09 & -0.21 & -0.10 & -0.13 & -0.08 & -0.14 & -0.09 \\
GesturePhaseSegmentationProcessed & \textbf{-0.07} & -0.13 & -0.19 & -0.17 & -0.15 & -0.13 & -0.17 & -0.15 \\
optdigits & \textbf{-0.01} & -0.01 & -0.02 & -0.05 & -0.02 & -0.05 & -0.04 & -0.05 \\
mfeat-factors & -0.06 & -0.04 & -0.05 & -0.03 & \textbf{-0.02} & -0.03 & -0.04 & -0.04 \\
mfeat-fourier & -0.08 & \textbf{-0.04} & -0.14 & -0.06 & -0.14 & -0.06 & -0.13 & -0.07 \\
qsar-biodeg & 0.01 & -0.02 & -0.01 & -0.03 & \textbf{0.01} & -0.03 & -0.03 & -0.03 \\
mfeat-zernike & \textbf{-0.05} & -0.07 & -0.15 & -0.09 & -0.07 & -0.05 & -0.14 & -0.08 \\
kr-vs-kp & -0.05 & \textbf{-0.02} & -0.18 & -0.10 & -0.18 & -0.10 & -0.08 & -0.05 \\
pc4 & -0.03 & -0.07 & -0.05 & -0.09 & -0.05 & \textbf{-0.02} & -0.06 & -0.06 \\
spambase & -0.04 & \textbf{-0.02} & -0.09 & -0.06 & -0.04 & -0.04 & -0.05 & -0.05 \\
satimage & -0.03 & -0.02 & -0.07 & -0.03 & -0.04 & \textbf{-0.01} & -0.03 & -0.02 \\
pc3 & \textbf{0.05} & 0.03 & 0.04 & 0.02 & -0.02 & 0.01 & 0.01 & 0.02 \\
texture & -0.10 & -0.07 & -0.06 & -0.08 & \textbf{-0.01} & -0.04 & -0.02 & -0.07 \\
mfeat-pixel & -0.04 & -0.00 & -0.05 & -0.00 & -0.00 & -0.00 & -0.04 & \textbf{-0.00} \\
splice & -1.36 & -0.49 & \textbf{-0.18} & -0.27 & -0.22 & -0.27 & -2.71 & -0.45 \\
first-order-theorem-proving & -0.08 & -0.11 & -0.17 & -0.10 & \textbf{-0.02} & -0.07 & -0.06 & -0.11 \\
    \bottomrule
\end{tabular}
\end{center}
\end{table*}
\end{landscape}

\begin{landscape}
\begin{table*}[h]
\caption{Baseline accuracy on datasets with pre-existing missingness}
\label{tbl:table14}
\centering
\begin{center}
\setlength\tabcolsep{1.5pt}
\begin{tabular}{lcccccccc}
    \toprule
    & \multicolumn{2}{c}{None} & \multicolumn{2}{c}{Simple} & \multicolumn{2}{c}{Iterative} & \multicolumn{2}{c}{MiceForest} \\
    & LSAM &   LightGBM & LSAM &  LightGBM & LSAM &  LightGBM & LSAM &   LightGBM \\
    \midrule
ipums\_la\_99-small & 0.66 & 0.81 & 0.67 & 0.81 & \textbf{0.66} & 0.81 & 0.66 & 0.77 \\
ipums\_la\_98-small & 0.69 & 0.86 & \textbf{0.68} & 0.86 & 0.69 & 0.86 & 0.69 & 0.85 \\
communities-and-crime-binary & 0.84 & 0.84 & 0.84 & 0.84 & \textbf{0.84} & 0.84 & 0.84 & 0.84 \\
jungle\_chess\_2pcs\_endgame\_rat\_panther & 1.00 & \textbf{0.99} & 1.00 & 0.99 & 1.00 & 0.99 & 1.00 & 0.99 \\
jungle\_chess\_2pcs\_endgame\_rat\_elephant & 1.00 & 0.99 & 0.99 & 0.99 & 0.99 & 0.99 & 0.99 & \textbf{0.99} \\
jungle\_chess\_2pcs\_endgame\_rat\_lion & 0.99 & 1.00 & 0.99 & 1.00 & \textbf{0.99} & 1.00 & 0.99 & 1.00 \\
kdd\_ipums\_la\_97-small & 0.98 & 0.98 & 0.98 & 0.98 & 0.98 & 0.98 & 0.97 & \textbf{0.77} \\
MiceProtein & 0.99 & 0.83 & 0.99 & 0.83 & 0.99 & \textbf{0.81} & 0.99 & 0.82 \\
cjs & 1.00 & 0.99 & 1.00 & 0.99 & 0.99 & 0.99 & 1.00 & \textbf{0.95} \\
SpeedDating & 0.80 & 0.81 & 0.80 & 0.81 & \textbf{0.78} & 0.82 & 0.80 & 0.82 \\
colleges\_usnews & 0.73 & 0.73 & 0.74 & 0.73 & 0.74 & 0.74 & 0.72 & \textbf{0.72} \\
    \bottomrule
\end{tabular}
\end{center}
\end{table*}
\end{landscape}

\begin{landscape}
\begin{table*}[h]
\caption{Baseline negative log-likelihood on datasets with pre-existing missingness}
\label{tbl:table15}
\centering
\begin{center}
\setlength\tabcolsep{1.5pt}
\begin{tabular}{lcccccccc}
    \toprule
    & \multicolumn{2}{c}{None} & \multicolumn{2}{c}{Simple} & \multicolumn{2}{c}{Iterative} & \multicolumn{2}{c}{MiceForest} \\
    & LSAM &   LightGBM & LSAM &  LightGBM & LSAM &  LightGBM & LSAM &   LightGBM \\
    \midrule
ipums\_la\_99-small & 5.03 & 0.71 & 4.97 & \textbf{0.71} & 5.03 & \textbf{0.71} & 5.01 & 0.81 \\
ipums\_la\_98-small & 0.88 & \textbf{0.60} & 0.91 & 0.60 & 0.88 & 0.60 & 0.89 & 0.66 \\
communities-and-crime-binary & 0.72 & \textbf{0.71} & 0.72 & 0.71 & 0.73 & 0.71 & 0.73 & 0.72 \\
jungle\_chess\_2pcs\_endgame\_rat\_panther & 0.02 & 0.04 & 0.02 & 0.04 & 0.02 & 0.04 & \textbf{0.02} & 0.04 \\
jungle\_chess\_2pcs\_endgame\_rat\_elephant & 0.02 & 0.03 & 0.02 & 0.03 & \textbf{0.02} & 0.03 & 0.03 & 0.05 \\
jungle\_chess\_2pcs\_endgame\_rat\_lion & 0.04 & \textbf{0.02} & 0.04 & 0.02 & 0.04 & 0.02 & 0.04 & 0.03 \\
kdd\_ipums\_la\_97-small & 0.15 & 0.17 & 0.15 & 0.17 & \textbf{0.15} & 0.17 & 0.22 & 1.29 \\
MiceProtein & 0.04 & 1.60 & \textbf{0.03} & 1.60 & 0.06 & 1.60 & 0.05 & 1.60 \\
cjs & \textbf{0.01} & 1.78 & 0.02 & 1.78 & 0.03 & 1.78 & 0.02 & 1.80 \\
SpeedDating & \textbf{0.83} & 0.97 & 0.85 & 0.97 & 0.89 & 0.96 & 0.84 & 0.96 \\
colleges\_usnews & 1.03 & 1.08 & 1.04 & 1.09 & \textbf{1.02} & 1.07 & 1.06 & 1.12 \\
    \bottomrule
\end{tabular}
\end{center}
\end{table*}
\end{landscape}

\hypertarget{appendix-d.}{%
\section*{Appendix D.}\label{appendix-d.}}
\addcontentsline{toc}{section}{Appendix D.}

Meta-learning was performed to study the characteristics of the datasets
where LSAM outperformed the comparator models. Random forest models were
trained to predict an LSAM win based on dataset characteristics under
several conditions. The first comparison looked at features importance
when LSAM beat LightGBM. Then feature importance was explored when LSAM
without imputation beat LSAM with imputation methods. The full feature
importance rankings for the complete but corrupted datasets and
incomplete datasets are presented below.

\begin{landscape}
\begin{table*}[h]
\caption{Meta-learning approach to determine feature importance with respect to LSAM outperforming benchmarks on corrupted datasets. Analysis is stratified by overall model type as well as out-of-the-box performance of LSAM compared to imputation.}
\label{tbl:table16}
\centering
\begin{center}
\setlength\tabcolsep{1.5pt}
\begin{tabular}{l|c|c|c|c|c|c|c|c}
    \toprule
  & \multicolumn{2}{r}{Overall: LightGBM vs LSAM} & \multicolumn{2}{r}{LSAM: Simple vs None} & \multicolumn{2}{r}{LSAM: Iterative vs None} & \multicolumn{2}{r}{LSAM: Miceforest vs None} \\
  Dataset Characteristic & Accuracy & NLL & Accuracy & NLL & Accuracy & NLL & Accuracy & NLL \\
    \midrule
NumberOfFeatures & 0.08 & 0.06 & \textbf{0.17} & 0.13 & \textbf{0.15} & 0.11 & \textbf{0.15} & 0.12 \\
NumberOfInstances & \textbf{0.21} & \textbf{0.22} & 0.11 & 0.10 & 0.11 & 0.12 & 0.12 & 0.13 \\
NumberOfClasses & 0.15 & 0.15 & 0.07 & 0.06 & 0.08 & 0.07 & 0.05 & 0.08 \\
NumberOfNumericFeatures & 0.12 & 0.15 & 0.10 & 0.12 & 0.11 & 0.11 & 0.10 & 0.11 \\
NumberOfSymbolicFeatures & 0.06 & 0.12 & 0.04 & 0.03 & 0.04 & 0.04 & 0.04 & 0.03 \\
NumericRatio & 0.16 & 0.12 & 0.10 & 0.11 & 0.09 & 0.11 & 0.11 & 0.10 \\
FeatureInstanceRatio & 0.12 & 0.11 & 0.15 & \textbf{0.13} & 0.13 & \textbf{0.16} & 0.14 & \textbf{0.14} \\
missingness\_MAR & 0.03 & 0.02 & 0.06 & 0.12 & 0.07 & 0.08 & 0.10 & 0.11 \\
missingness\_MCAR & 0.03 & 0.03 & 0.07 & 0.09 & 0.09 & 0.10 & 0.09 & 0.10 \\
missingness\_MNAR & 0.05 & 0.03 & 0.13 & 0.11 & 0.12 & 0.10 & 0.10 & 0.10 \\
  \bottomrule
\end{tabular}
\end{center}
\end{table*}
\end{landscape}

\begin{landscape}
\begin{table*}[h]
\caption{Meta-learning approach to determine feature importance with respect to LSAM outperforming benchmarks on datasets with pre-existing missingness. Analysis is stratified by overall model type as well as out-of-the-box performance of LSAM compared to imputation.}
\label{tbl:table16}
\centering
\begin{center}
\setlength\tabcolsep{1.5pt}
\begin{tabular}{l|c|c|c|c|c|c|c|c}
    \toprule
  & \multicolumn{2}{r}{Overall: LightGBM vs LSAM} & \multicolumn{2}{r}{LSAM: Simple vs None} & \multicolumn{2}{r}{LSAM: Iterative vs None} & \multicolumn{2}{r}{LSAM: Miceforest vs None} \\
  Dataset Characteristic & Accuracy & NLL & Accuracy & NLL & Accuracy & NLL & Accuracy & NLL \\
    \midrule
NumberOfFeatures & 0.09 & 0.11 & \textbf{0.16} & 0.04 & \textbf{0.19} & \textbf{0.26} & \textbf{0.22} & 0.08 \\
NumberOfInstances & 0.23 & 0.07 & 0.11 & \textbf{0.24} & 0.11 & 0.13 & 0.12 & 0.15 \\
NumberOfClasses & 0.11 & 0.03 & 0.09 & 0.08 & 0.02 & 0.06 & 0.04 & 0.05 \\
NumberOfNumericFeatures & 0.09 & \textbf{0.27} & 0.10 & 0.16 & 0.14 & 0.09 & 0.17 & 0.18 \\
NumberOfSymbolicFeatures & \textbf{0.25} & 0.11 & 0.13 & 0.06 & 0.13 & 0.10 & 0.09 & 0.06 \\
NumericRatio & 0.09 & 0.23 & 0.16 & 0.17 & 0.18 & 0.12 & 0.07 & 0.11 \\
FeatureInstanceRatio & 0.06 & 0.14 & 0.11 & 0.21 & 0.07 & 0.08 & 0.15 & \textbf{0.23} \\
FractionMissingValues & 0.09 & 0.05 & 0.14 & 0.04 & 0.17 & 0.16 & 0.15 & 0.14 \\
  \bottomrule
\end{tabular}
\end{center}
\end{table*}
\end{landscape}

\newpage

\hypertarget{references}{%
\section*{References}\label{references}}
\addcontentsline{toc}{section}{References}

\hypertarget{refs}{}
\begin{CSLReferences}{1}{0}
\leavevmode\vadjust pre{\hypertarget{ref-Abiri_2019}{}}%
Abiri N, Linse B, Edén P, Ohlsson M. Establishing strong imputation
performance of a denoising autoencoder in a wide range of missing data
problems. Neurocomputing {[}Internet{]}. 2019 Nov;365:137--46. Available
from: \url{http://dx.doi.org/10.1016/j.neucom.2019.07.065}

\leavevmode\vadjust pre{\hypertarget{ref-bertsimas2021prediction}{}}%
Bertsimas D, Delarue A, Pauphilet J. Prediction with missing data
{[}Internet{]}. 2021. Available from:
\url{https://arxiv.org/abs/2104.03158}

\leavevmode\vadjust pre{\hypertarget{ref-JMLR:v18:17-073}{}}%
Bertsimas D, Pawlowski C, Zhuo YD. From predictive methods to missing
data imputation: An optimization approach. Journal of Machine Learning
Research {[}Internet{]}. 2018;18(196):1--39. Available from:
\url{http://jmlr.org/papers/v18/17-073.html}

\leavevmode\vadjust pre{\hypertarget{ref-wandb}{}}%
Biewald L. Experiment tracking with weights and biases {[}Internet{]}.
2020. Available from: \url{https://www.wandb.com/}

\leavevmode\vadjust pre{\hypertarget{ref-boudiaf2020unifying}{}}%
Boudiaf M, Rony J, Ziko IM, Granger E, Pedersoli M, Piantanida P, et al.
A unifying mutual information view of metric learning: Cross-entropy vs.
Pairwise losses. In: Computer vision {\textendash} {ECCV} 2020
{[}Internet{]}. Springer International Publishing; 2020. p. 548--64.
Available from: \url{https://doi.org/10.1007/978-3-030-58539-6_33}

\leavevmode\vadjust pre{\hypertarget{ref-jax2018github}{}}%
Bradbury J, Frostig R, Hawkins P, Johnson MJ, Leary C, Maclaurin D, et
al. {JAX}: Composable transformations of {P}ython+{N}um{P}y programs
{[}Internet{]}. 2018. Available from: \url{http://github.com/google/jax}

\leavevmode\vadjust pre{\hypertarget{ref-JSSv045i03}{}}%
Buuren S van, Groothuis-Oudshoorn K. Mice: Multivariate imputation by
chained equations in r. Journal of Statistical Software, Articles
{[}Internet{]}. 2011;45(3):1--67. Available from:
\url{https://www.jstatsoft.org/v045/i03}

\leavevmode\vadjust pre{\hypertarget{ref-chang2018dropout}{}}%
Chang C-H, Rampasek L, Goldenberg A. Dropout feature ranking for deep
learning models {[}Internet{]}. 2018. Available from:
\url{https://arxiv.org/abs/1712.08645}

\leavevmode\vadjust pre{\hypertarget{ref-davey2020systematic}{}}%
Davey A, Dai T. A systematic approach to identify and evaluate missing
data patterns and mechanisms in multivariate educational, social, and
behavioral research {[}Internet{]}. 2020. Available from:
\url{https://arxiv.org/abs/2007.14296}

\leavevmode\vadjust pre{\hypertarget{ref-Janez2006}{}}%
Demšar J. Statistical comparisons of classifiers over multiple data
sets. J Mach Learn Res. 2006 Dec;7:1--30.

\leavevmode\vadjust pre{\hypertarget{ref-decisionTreeMissing}{}}%
Gavankar S, Sawarkar S.
\href{https://doi.org/10.1109/AIMS.2015.29}{Decision tree: Review of
techniques for missing values at training, testing and compatibility}.
In: 2015 3rd international conference on artificial intelligence,
modelling and simulation (AIMS). 2015. p. 122--6.

\leavevmode\vadjust pre{\hypertarget{ref-Jeong2022}{}}%
Jeong H, Wang H, Calmon FP. Fairness without imputation: A decision tree
approach for fair prediction with missing values. Proceedings of the
{AAAI} Conference on Artificial Intelligence {[}Internet{]}. 2022
Jun;36(9):9558--66. Available from:
\url{https://doi.org/10.1609/aaai.v36i9.21189}

\leavevmode\vadjust pre{\hypertarget{ref-ke2017lightgbm}{}}%
Ke G, Meng Q, Finley T, Wang T, Chen W, Ma W, et al. Lightgbm: A highly
efficient gradient boosting decision tree. Advances in neural
information processing systems. 2017;30:3146--54.

\leavevmode\vadjust pre{\hypertarget{ref-kingma2017adam}{}}%
Kingma DP, Ba J. Adam: A method for stochastic optimization
{[}Internet{]}. 2017. Available from:
\url{https://arxiv.org/abs/1412.6980}

\leavevmode\vadjust pre{\hypertarget{ref-informMissingLi}{}}%
Li J, Wang M, Steinbach MS, Kumar V, Simon GJ.
\href{https://doi.org/10.1109/ICBK.2018.00062}{Don't do imputation:
Dealing with informative missing values in EHR data analysis}. In: Soon
OY, Chen H, Wu X, Aggarwal C, editors. Proceedings - 9th IEEE
international conference on big knowledge, ICBK 2018. Institute of
Electrical; Electronics Engineers Inc.; 2018. p. 415--22. (Proceedings -
9th IEEE international conference on big knowledge, ICBK 2018).

\leavevmode\vadjust pre{\hypertarget{ref-little2002statistical}{}}%
Little RJA, Rubin DB. Statistical analysis with missing data
{[}Internet{]}. Wiley; 2002. (Wiley series in probability and
mathematical statistics. Probability and mathematical statistics).
Available from: \url{http://books.google.com/books?id=aYPwAAAAMAAJ}

\leavevmode\vadjust pre{\hypertarget{ref-ma2019eddi}{}}%
Ma C, Tschiatschek S, Palla K, Hernandez Lobato JM, Nowozin S, Zhang C.
EDDI: Efficient dynamic discovery of high-value information with partial
VAE. In: International conference on machine learning {[}Internet{]}.
2019. Available from: \url{https://tinyurl.com/d4uu9acx}

\leavevmode\vadjust pre{\hypertarget{ref-scikit-learn}{}}%
Pedregosa F, Varoquaux G, Gramfort A, Michel V, Thirion B, Grisel O, et
al. Scikit-learn: Machine learning in {P}ython. Journal of Machine
Learning Research. 2011;12:2825--30.

\leavevmode\vadjust pre{\hypertarget{ref-rubin:1987}{}}%
Rubin DB. Multiple imputation for nonresponse in surveys. Wiley; 1987.
p. 258.

\leavevmode\vadjust pre{\hypertarget{ref-Stekhoven2011}{}}%
Stekhoven DJ, Buhlmann P. MissForest--non-parametric missing value
imputation for mixed-type data. Bioinformatics {[}Internet{]}. 2011
Oct;28(1):112--8. Available from:
\url{http://dx.doi.org/10.1093/bioinformatics/btr597}

\leavevmode\vadjust pre{\hypertarget{ref-OpenML2013}{}}%
Vanschoren J, Rijn JN van, Bischl B, Torgo L. OpenML: Networked science
in machine learning. SIGKDD Explorations {[}Internet{]}.
2013;15(2):49--60. Available from:
\url{http://doi.acm.org/10.1145/2641190.2641198}

\leavevmode\vadjust pre{\hypertarget{ref-vaswani2017attention}{}}%
Vaswani A, Shazeer N, Parmar N, Uszkoreit J, Jones L, Gomez AN, et al.
Attention is all you need. In: Guyon I, Luxburg UV, Bengio S, Wallach H,
Fergus R, Vishwanathan S, et al., editors. Advances in neural
information processing systems {[}Internet{]}. Curran Associates, Inc.;
2017. Available from:
\url{https://proceedings.neurips.cc/paper/2017/file/3f5ee243547dee91fbd053c1c4a845aa-Paper.pdf}

\leavevmode\vadjust pre{\hypertarget{ref-DBLP:journalsux2fcorrux2fabs-2010-07468}{}}%
Zhuang J, Tang T, Ding Y, Tatikonda S, Dvornek NC, Papademetris X, et
al. AdaBelief optimizer: Adapting stepsizes by the belief in observed
gradients. CoRR {[}Internet{]}. 2020;abs/2010.07468. Available from:
\url{https://arxiv.org/abs/2010.07468}

\end{CSLReferences}

\end{document}